%% file: iclr2025_conference.tex
\documentclass{article} 
\usepackage{iclr2025_conference,times}

\input{math_commands.tex}

\usepackage{hyperref}
\usepackage{url}
\usepackage{graphicx}
\usepackage{array}       
\usepackage{booktabs}    
\usepackage{graphicx}    
\usepackage{amsmath}     
\usepackage{caption}    
\usepackage{adjustbox}
\usepackage{multirow}
\usepackage{subcaption}
\usepackage{amssymb}
\usepackage{colortbl}
\usepackage{wrapfig}

\newcommand{\ours}{Finedefics}

\title{Analyzing and Boosting the Power of Fine-Grained Visual Recognition for Multi-modal Large Language Models}

\author{Hulingxiao He\textsuperscript{\rm 1}, Geng Li\textsuperscript{\rm 1}, Zijun Geng\textsuperscript{\rm 1}, Jinglin Xu\textsuperscript{\rm 2}, Yuxin Peng\textsuperscript{\rm 1}\thanks{Corresponding author.} \\
\textsuperscript{\rm 1}Wangxuan Institute of Computer Technology, Peking University\\
\textsuperscript{\rm 2}School of Intelligence Science and Technology, University of Science and Technology Beijing\\
\texttt{\{hehulingxiao,ligeng\}@stu.pku.edu.cn},
\texttt{gengzijun2024@163.com} \\
\texttt{xujinglinlove@gmail.com},
\texttt{pengyuxin@pku.edu.cn} 
}

\iclrfinalcopy 
\begin{document}

\maketitle
\begin{abstract}
Multi-modal large language models (MLLMs) have shown remarkable abilities in various visual understanding tasks. However, MLLMs still struggle with fine-grained visual recognition (FGVR), which aims to identify subordinate-level categories from images. This can negatively impact more advanced capabilities of MLLMs, such as object-centric visual question answering and reasoning. In our study, we revisit three quintessential capabilities of MLLMs for FGVR, including object information extraction, category knowledge reserve, object-category alignment, and position of the root cause as a misalignment problem. To address this issue, we present \textbf{\ours}, an MLLM that enhances the model's FGVR capability by incorporating informative attribute descriptions of objects into the training phase. We employ contrastive learning on object-attribute pairs and attribute-category pairs simultaneously and use examples from similar but incorrect categories as hard negatives, naturally bringing representations of visual objects and category names closer. Extensive evaluations across multiple popular FGVR datasets demonstrate that \ours~outperforms existing MLLMs of comparable parameter sizes, showcasing its remarkable efficacy. The code is available at \href{https://github.com/PKU-ICST-MIPL/Finedefics_ICLR2025}{\texttt{https://github.com/PKU-ICST-MIPL/Finedefics\_ICLR2025}}. 
\end{abstract}

\section{Introduction}

Multi-modal Large Language Models (MLLMs) \citep{bai2023qwen, chen2023shikra, zhang2023internlm, zhu2023minigpt, dong2024internlm, liu2024visual, liu2024improved, laurenccon2024obelics, laurenccon2024matters} have achieved remarkable advancements in understanding visual data, showcasing potential in advancing general artificial intelligence. These models enable users to interact with images as inputs, fostering seamless communication grounded in visual information. The impressive capabilities allow MLLMs to excel in various vision tasks while adeptly handling complex content comprehension and generation. However, despite their versatility and linguistic proficiency, MLLMs still face challenges in a fundamental task of machine vision: fine-grained visual recognition (FGVR) \citep{zhang2024visually, geigle2024african}, which aims at identifying subordinate-level categories, such as specific species of animals or plants \citep{wei2021fine}. Poor FGVR performance of MLLMs hinders them from performing more advanced tasks like object-centric visual question answering and reasoning \citep{zhang2024visually}. For example, in smart agriculture, poor FGVR performance of pests may lead to incorrect treatment strategies and large-scale reduction in food production. 

\begin{figure*}[t]
    \centering
    \includegraphics[width=0.98\linewidth]{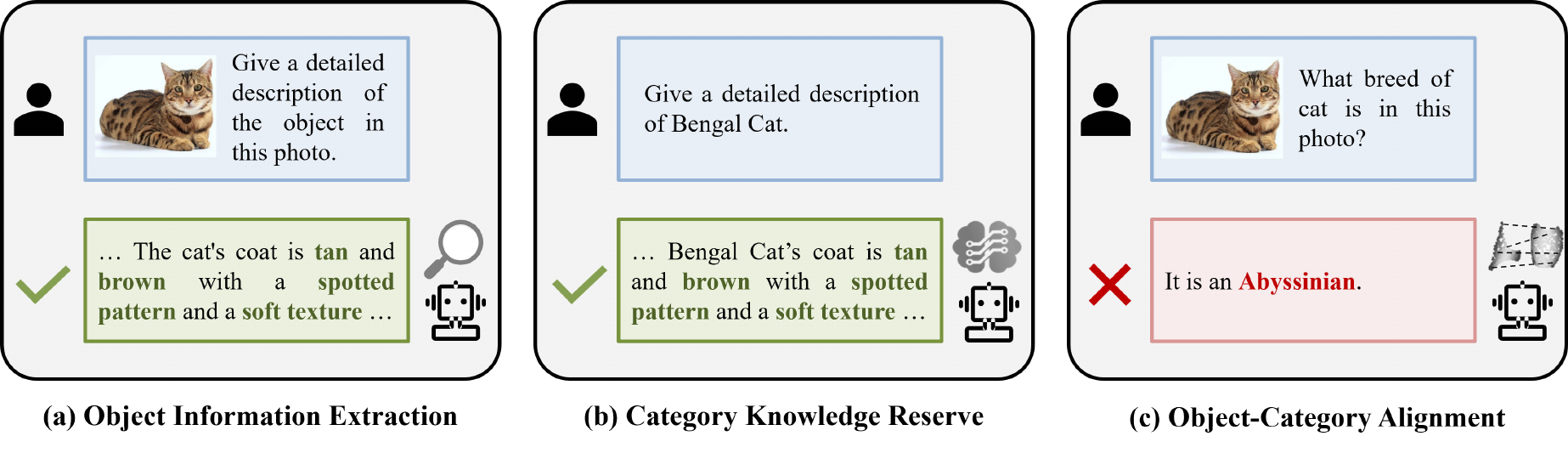}
    \caption{Three quintessential capabilities of MLLMs for fine-grained visual recognition. Current MLLMs possess acceptable capabilities in image information extraction and category knowledge reserve but struggle with aligning objects to their corresponding subordinate-level categories.}
    \label{fig:motivation}
\end{figure*}

Early works have investigated the phenomenon \citep{zhang2024visually} and attempted to improve the FGVR performance of MLLMs by integrating open-set classification data into pre-training or fine-tuning stage \citep{geigle2024african, zhang2024visually}. However, fine-tuning solely on the classification task harms the general capability of the instruction following, while purely integrating classification-focused data into the instruction tuning data brings limited improvement \citep{geigle2024african}, making their direct utilization impractical. To understand why MLLMs underperform in FGVR, we revisit and evaluate three quintessential capabilities of MLLMs as shown in Figure \ref{fig:motivation}: (a) \textbf{Object information extraction}. It is essential to accurately and fully extract the necessary information for distinguishing objects. (b) \textbf{Category knowledge reserve}. MLLMs should reserve sufficient knowledge of subordinate-level categories. (c) \textbf{Object-category alignment}. With rich visual information extracted and sufficient category knowledge reserved, visual objects and category names should be aligned in the representation space to enhance classification performance.

We analyze the representation space of MLLMs and their corresponding visual language models (VLMs) like CLIP \citep{radford2021learning}, revealing that: (1) \textit{Object information lost exists between VLMs and MLLMs but is not the bottleneck.} During the propagation of object features output from the vision encoder in modality connector and language model layers, the necessary visual information for distinguishing objects is almost preserved. Our observation is consistent with object hallucination in MLLMs \citep{zhou2023analyzing}, which finds that the modality connector tends to decrease the representation discriminability, while LLM has no significant impact. (2) \textit{Category knowledge is relatively sufficient, but category names cannot fully capture the semantics.} Existing LLMs utilized in MLLMs can output detailed and distinguishing descriptions about subordinate-level categories, but category names are not discriminative in the representation space of LLMs. (3) \textit{Misalignment between the visual object and category name leads to underperformance.} Despite a slight reduction in both object and category representation discriminability, the current learned modality connector is insufficient for effectively matching visual object representations to subordinate-level category names, as most VLMs do.

Motivated by the aforementioned analysis, we propose \textbf{\ours}, an MLLM designed to enhance the model's ability to identify subordinate-level visual object categories. Our framework builds upon Idefics2 \citep{laurenccon2024matters} and is specifically tailored to boost the power of FGVR. To facilitate alignment between visual objects and category names, descriptions summarized from information visual attributes are utilized as the intermediate point to bind them in the representation space of LLMs. Concretely, we separately feed token sequences of visual objects, attribute descriptions, and category names into MLLMs to obtain global representations from the last layer of LLMs respectively. Contrastive learning is performed on global representations of object-attribute pairs and attribute-category pairs simultaneously, with additional hard negatives from similar but incorrect categories, bringing representations of visual objects and category names closer. Subsequently, being trained solely on both open-set and closed-set FGVR data through instruction tuning, \ours~demonstrates exceptional FGVR performance gains. Benefiting from attribute augmented alignment, \ours~outperforms established counterparts across six popular FGVR datasets and notably surpassing Idefics2 \citep{laurenccon2024matters} and Qwen-VL-Chat \citep{bai2023qwen} by an average of +10.89\% and +9.43\%, respectively.

In summary, our contributions are as follows: (i) We revisit the quintessential capabilities of MLLMs for FGVR and investigate the root cause of underperformance in FGVR: misalignment between visual objects and category names. (ii) We propose \ours~for enhancing the model's FGVR accuracy, which uses informative attribute descriptions to effectively align visual objects and category names in the representation space of LLMs. (iii) With extensive experiments on six popular FGVR datasets, we demonstrate the superiority of \ours.

\begin{figure*}[t]
 \vspace{-2ex}
  \centering
  \begin{subfigure}[b]{0.32\linewidth}
    \includegraphics[width=\textwidth]{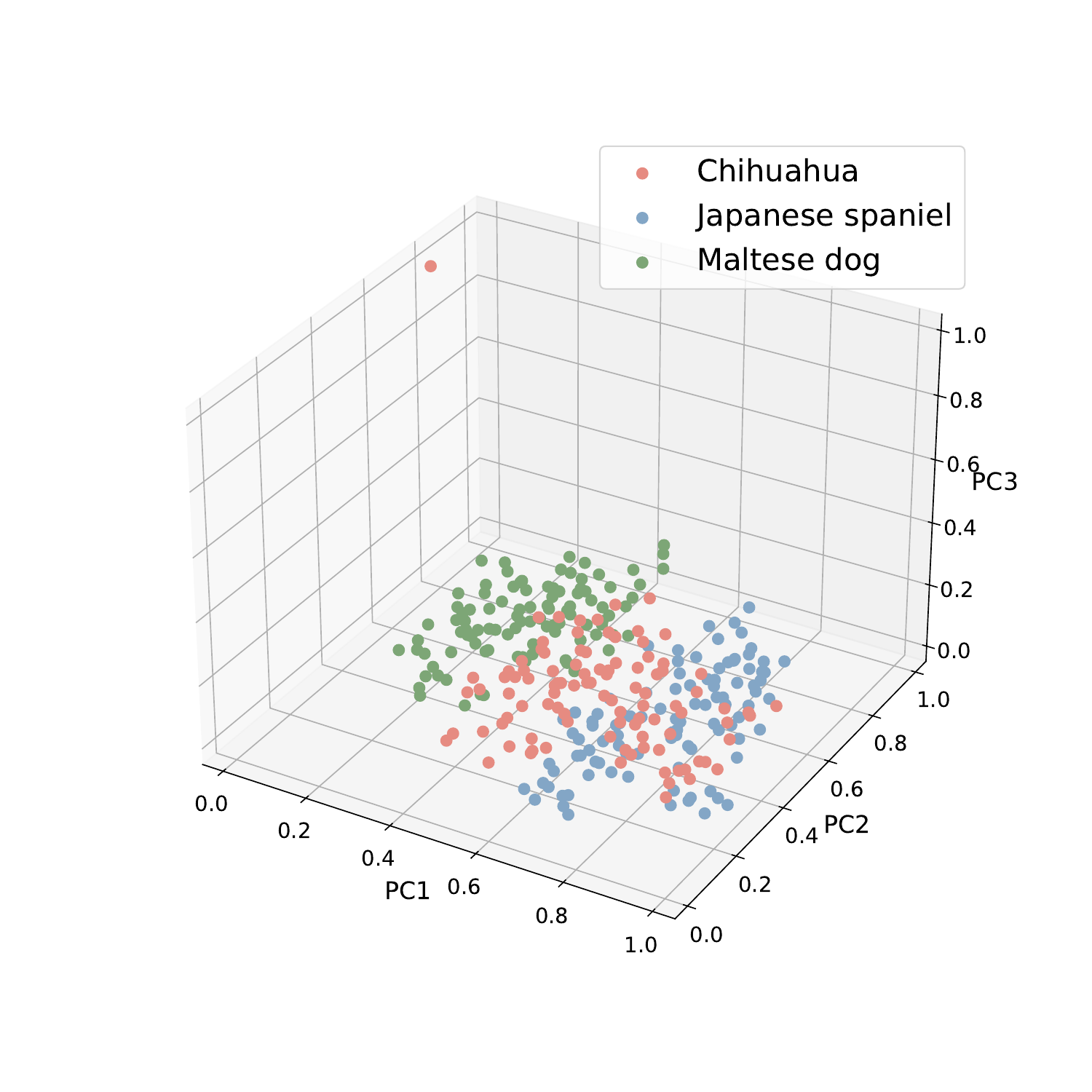}
    \caption{Object dist. of SigLIP.}
    \label{fig:vis_img_siglip}
  \end{subfigure}
  \hfill
  \begin{subfigure}[b]{0.32\linewidth}
    \includegraphics[width=\textwidth]{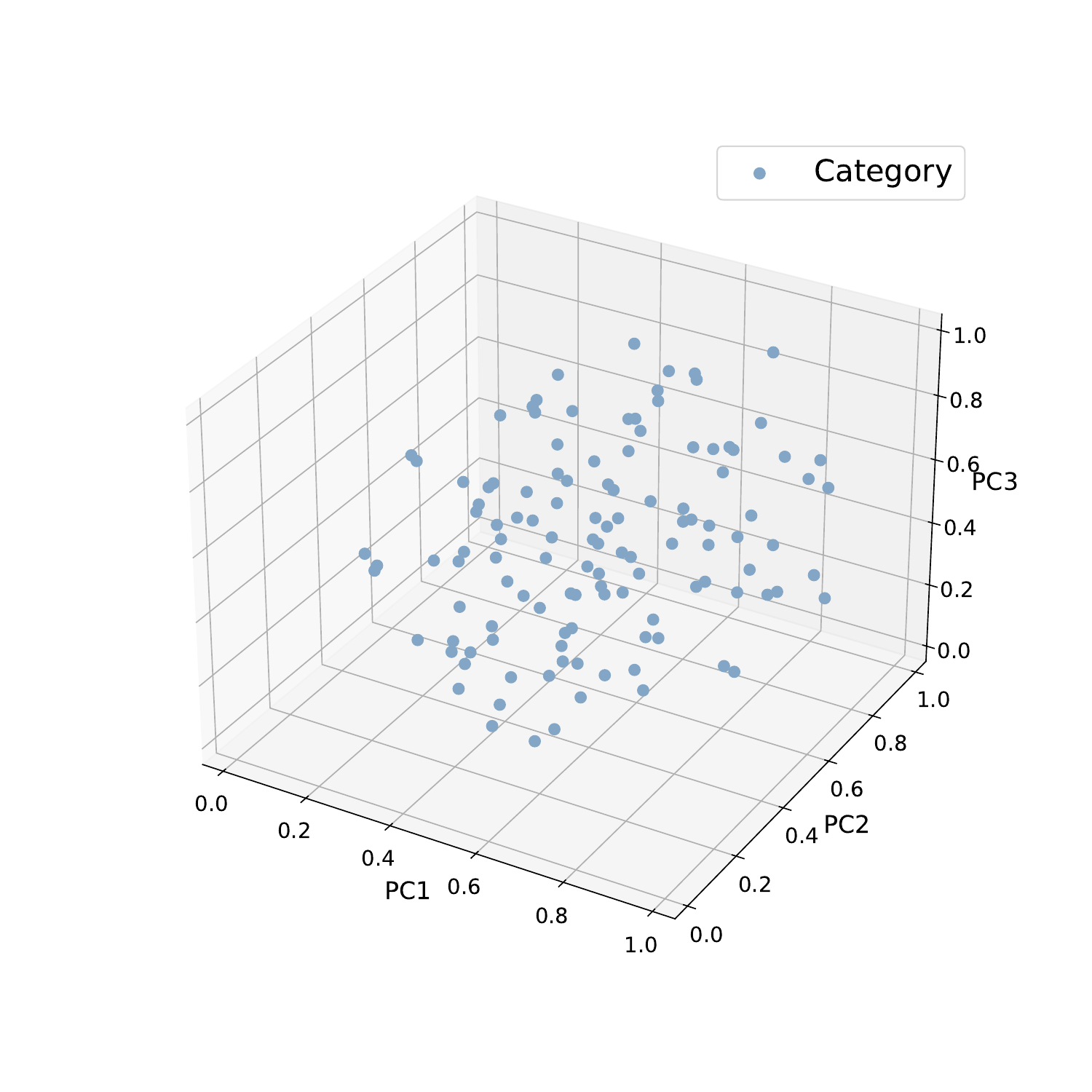}
    \caption{Category dist. of SigLIP.}
    \label{fig:vis_txt_siglip}
  \end{subfigure}
  \hfill
  \begin{subfigure}[b]{0.32\linewidth}
    \includegraphics[width=\textwidth]{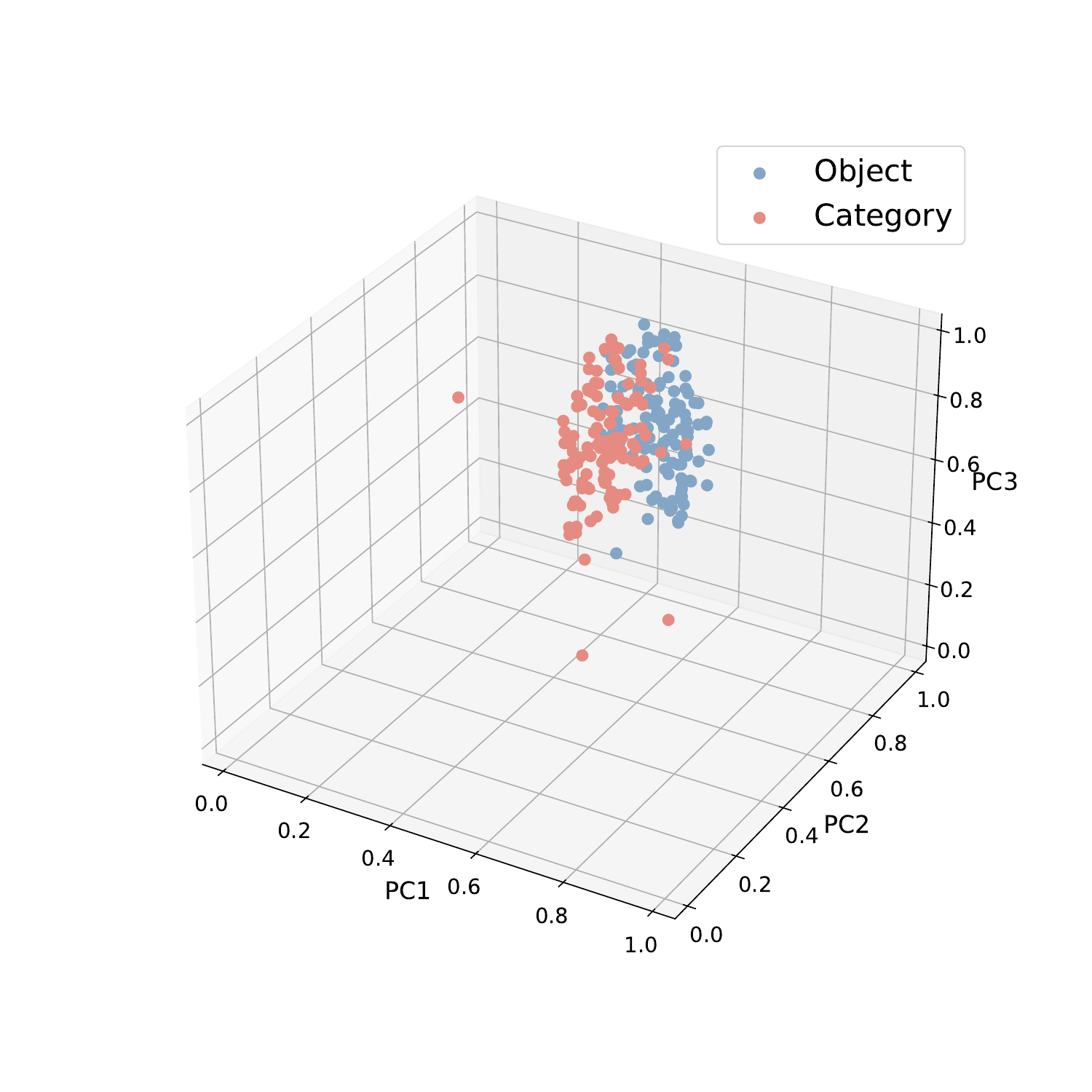}
    \caption{Object-category dist. of SigLIP.}
    \label{fig:vis_img_txt_siglip}
  \end{subfigure}

  \begin{subfigure}[b]{0.32\linewidth}
    \includegraphics[width=\textwidth]{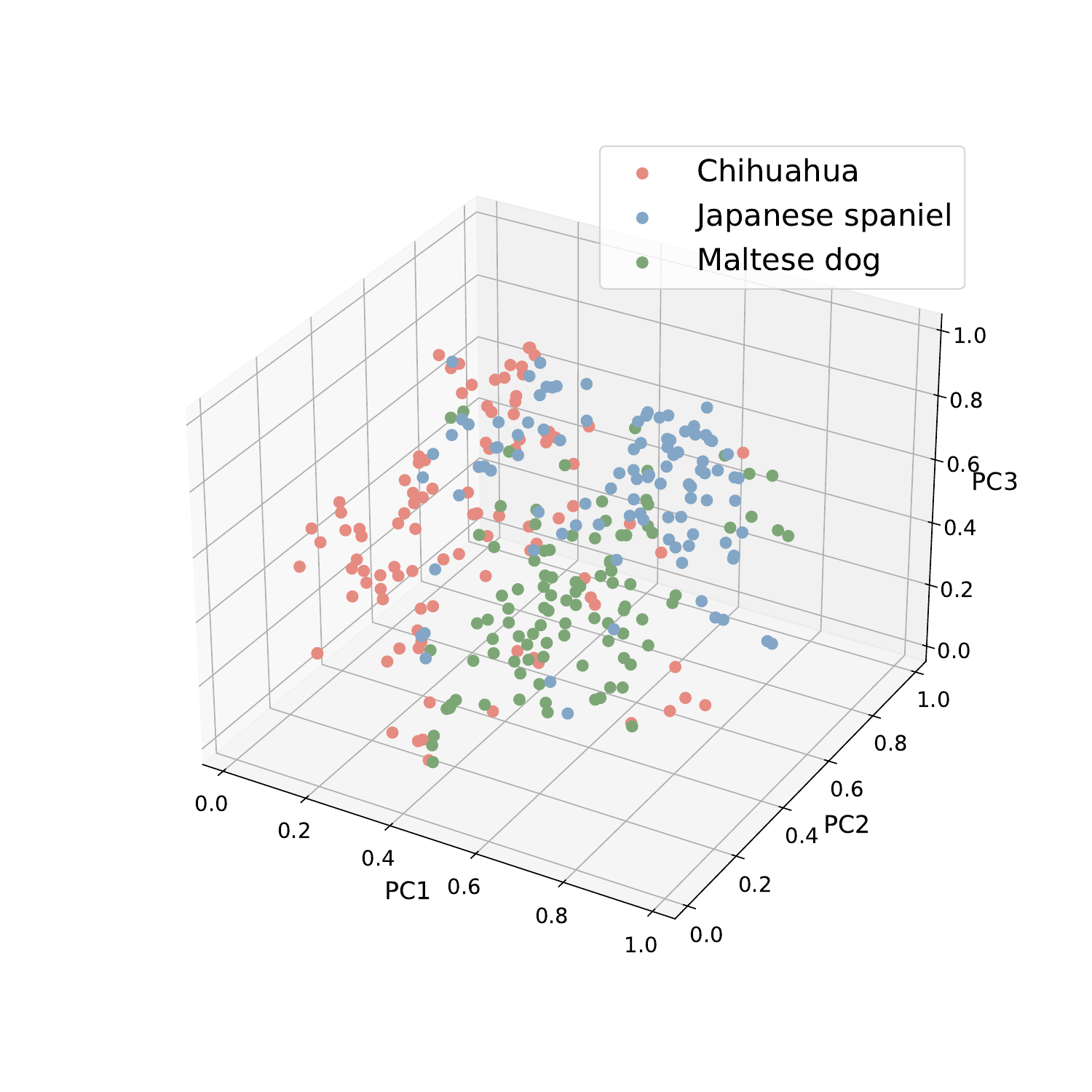}
    \caption{Object dist. of Idefics2.}
    \label{fig:vis_img_Idefics2}
  \end{subfigure}
  \hfill
  \begin{subfigure}[b]{0.32\linewidth}
    \includegraphics[width=\textwidth]{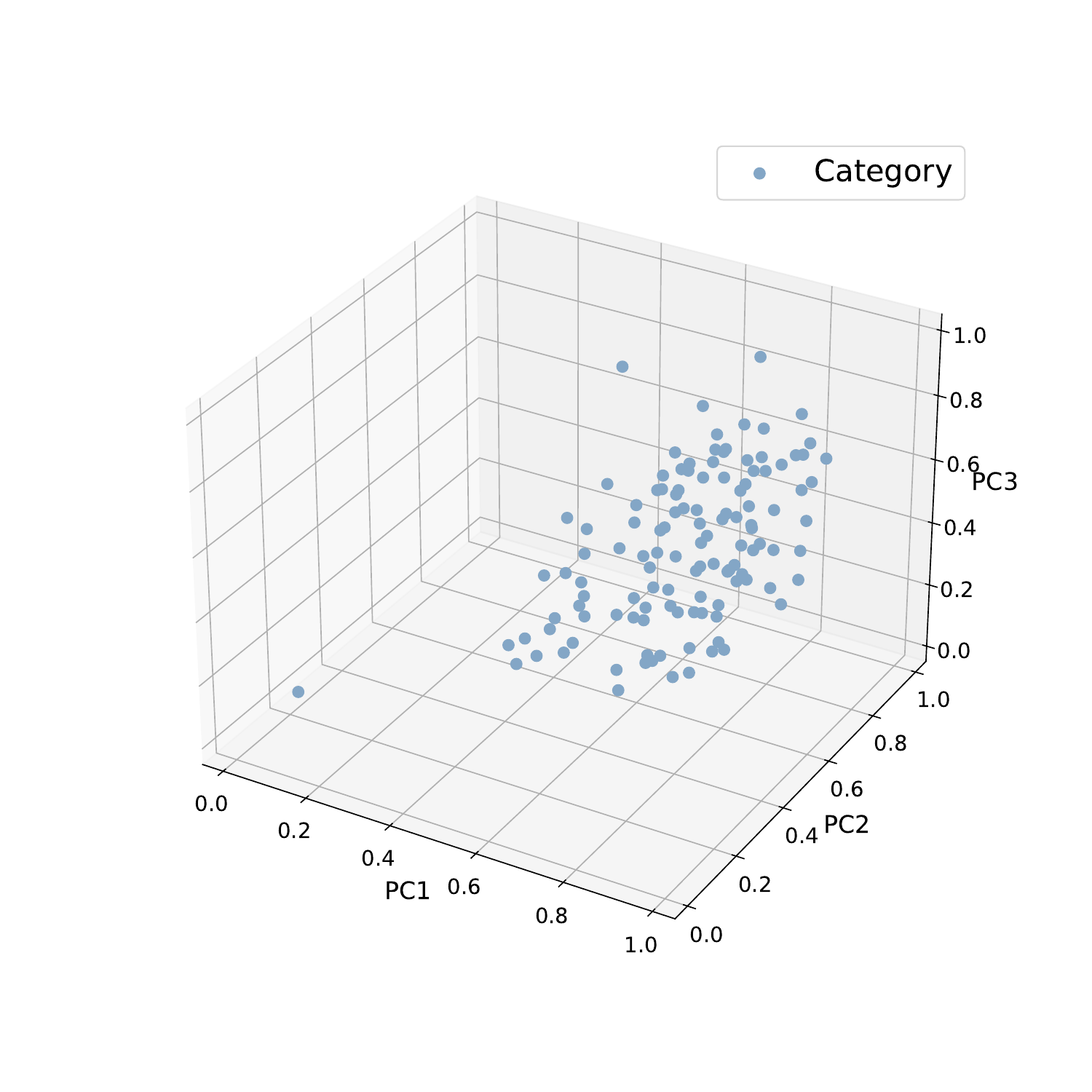}
    \caption{Category dist. of Idefics2.}
    \label{fig:vis_txt_Idefics2}
  \end{subfigure}
  \hfill
  \begin{subfigure}[b]{0.33\linewidth}
    \includegraphics[width=\textwidth]{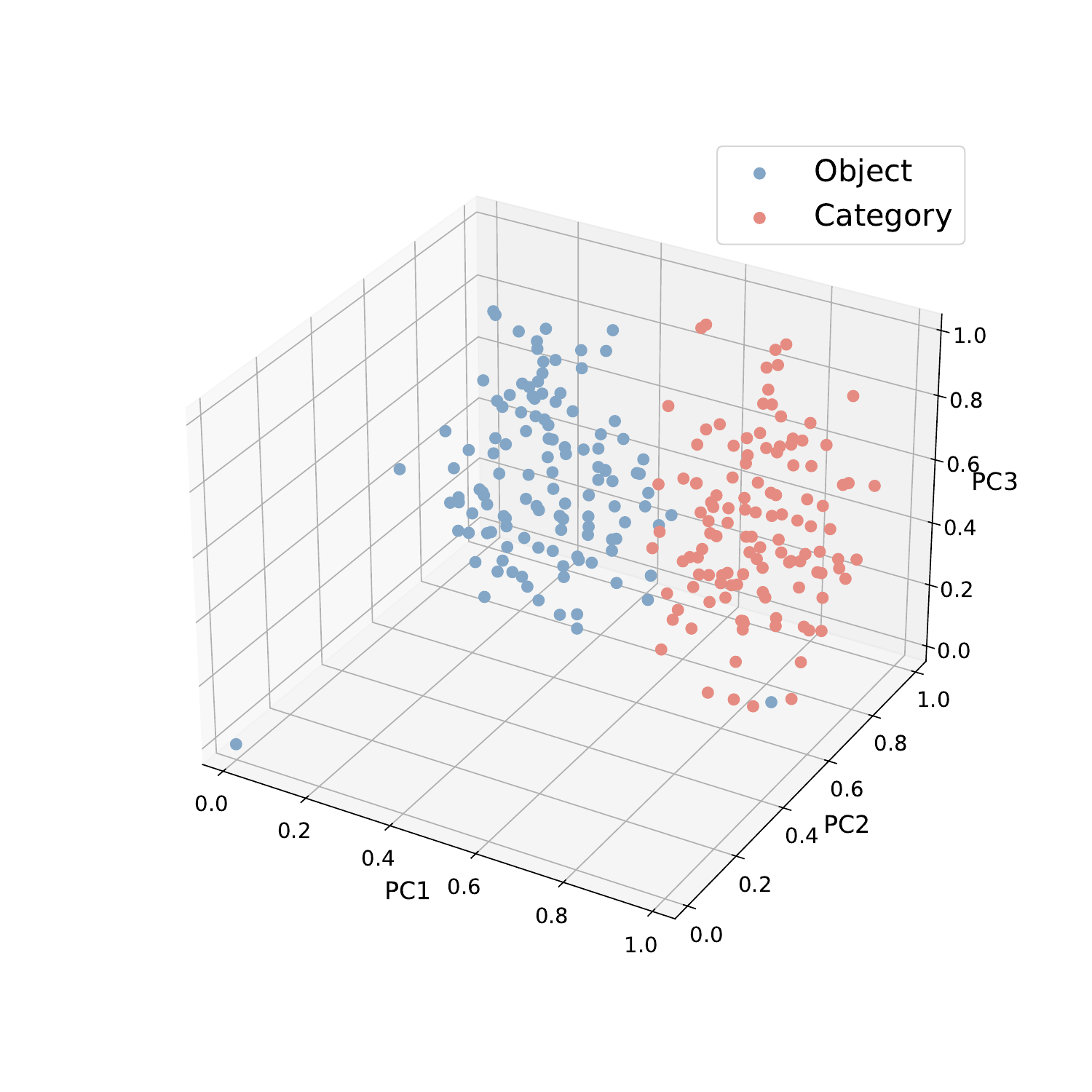}
    \caption{Object-category dist. of Idefics2.}
    \label{fig:vis_img_txt_Idefics2}
  \end{subfigure}

  \vspace{2ex}
  \caption{Object/Category/Object-category representation visualization of SigLIP and Idefics2.}
  \label{fig:vis}
\end{figure*}

\section{Why Do Multi-modal Large Language Models Underperform in Fine-Grained Visual Recognition?}

This section scrutinizes the root causes of underperformance in FGVR via comprehensive empirical analyses on three quintessential capabilities of MLLMs: \textit{object information extraction}, \textit{category knowledge reserve}, and \textit{object-category alignment} by comparing the representation space with corresponding VLMs (Idefics2 \citep{laurenccon2024obelics} and SigLIP \citep{zhai2023sigmoid} in our experiments). 

\textbf{Notations.} Assuming an image $I_{i}$ containing an object $O_{i}$ is processed by the vision encoder $\mathbf{V}_{\alpha}$ and learnable modality connector $\mathbf{F}_{\beta}$ to be transformed into a visual object token sequence of length $m$: $S_{o}^{i}=[o_{1}^{i}, o_{2}^{i}, \dots, o_{m}^{i}]$. Input category name in textual modality $C_{i}$ is passed through an embedding layer $E_{\phi}$ of the LLM to obtain the category embedding sequence of length $n$: $S_{c}^{i}=[c_{1}^{i}, c_{2}^{i}, \dots, c_{n}^{i}]$. Subsequently, the object embedding sequence $S_{o}^{i}$ and category embedding sequence $S_{c}^{i}$ are individually passed through the LLM layers $\mathbf{L}_{\theta}$ to obtain the output from the last layer:
\begin{subequations}
\begin{align}
    H_{o}^{i} = \mathbf{L}_{\theta} (S_{o}^{i}),    
\end{align}
\begin{align}
    H_{c}^{i} = \mathbf{L}_{\theta} (S_{c}^{i}),   
\end{align}
\end{subequations}
where $H_{o}^{i}=[\hat{o}_{1}^{i}, \hat{o}_{2}^{i}, \dots, \hat{o}_{m}^{i}]$, and $H_{c}^{i}(l)=[\hat{c}_{1}^{i}, \hat{c}_{2}^{i}, \dots, \hat{c}_{n}^{i}]$. Afterward, we select two ways to represent the global semantics of output sequence following \citep{zhang2024visually}: 1) last token embedding $\hat{o}_{m}^{i}$, $\hat{c}_{n}^{i}$, and (b) average of the token embedding sequence $\bar{o}^{i}=(\sum_{k=1}^{m}\hat{o}_{k}^{i})/m$, $\bar{c}^{i}=(\sum_{k=1}^{n}\hat{c}_{k}^{i})/n$. For VLMs, the projected \texttt{[CLS]} embedding outputs from last layer of vision encoder $\textbf{V}_{\alpha}$ and text encoder $\textbf{T}_{\gamma}$ are taken to represent the global semantics of $O_{i}$ and $C_{i}$, denoted as $\hat{o}_{\text{CLS}}^{i}$ and $\hat{c}_{\text{CLS}}^{i}$, respectively.

\subsection{Object information extraction} 
In the task of FGVR, a model that excels at object information extraction is required to have discriminative representations, i.e., large inter-class distance and small intra-class variance. To compare the object representation space of SigLIP and Idefics2, we select three subordinate-level categories \texttt{["Chihuahua","Japanese spaniel","Maltese dog"]} from Stanford Dog-120 \citep{krause20133d}, and randomly sample 100 examples per category for t-SNE \citep{van2008visualizing} visualization. As shown in Figure \ref{fig:vis_img_siglip}, since the vision encoder $\mathbf{V}_{\alpha}$ is normally frozen throughout training \citep{liu2024visual,liu2024improved}, the output object token sequence $S_{o}^{i}$ preserves discriminative information for classification. We then hypothesize that the information is lost after propagating through the modality connector $\mathbf{F}_{\beta}$ and LLM layers $\mathbf{L}_{\theta}$. However, various objects belonging to the same subordinate-level categories can still cluster together and distance from each other, as illustrated in Figure \ref{fig:vis_img_Idefics2}. To quantitatively compare the representation discriminability, we use feature probing experiments \citep{zhang2024visually} to test the hypothesis. Concretely, on top of the last token embedding $\hat{o}_{m}^{i}$ or average of the token embedding sequence $\bar{o}^{i}$, we train a linear classifier on the training set of Oxford-IIIT Pet-37 \citep{parkhi2012cats} and evaluate on the test set. In Table \ref{tab:probe_visual}, we observe that although information is lost, the impact on the performance is limited.

\begin{table*}[t]
  \caption{
    Feature probing on Idefics2 and SigLIP with features of objects and category descriptions.
  }  
  \label{tab:probe}
  \centering
  \subcaptionbox{Object features.\label{tab:probe_visual}}{
    \centering
    \resizebox{0.36\textwidth}{!}{%
    \begin{tabular}{@{}l c| c|c@{}}
    	\toprule[0.15em]
     
    	 \textbf{Model}           && \textbf{Feature Type} & \textbf{Acc.}  \\
    	\midrule
            \multirow{2}{*}{Idefics2} && Last             &   94.99            \\
                                         && Avg.             &   90.24            \\
            \midrule
            \multirow{2}{*}{SigLIP}      && CLS              &   95.28            \\
                                         && Avg.             &   94.44            \\ 
    	\bottomrule[0.15em]
    \end{tabular}
    }
  }
  \hspace{3em}
  \subcaptionbox{Category description features.\label{tab:probe_textual}}{
  \resizebox{0.36\textwidth}{!}{%
    \begin{tabular}{@{}l c| c|c@{}}
    	\toprule[0.15em]
    	 \textbf{Model}           && \textbf{Feature Type} & \textbf{Acc.}  \\
    	\midrule
            \multirow{2}{*}{Idefics2} && Last             &   92.51           \\
                                         && Avg.             &   90.41           \\
            \midrule
            \multirow{2}{*}{SigLIP}      && CLS              &   84.70            \\
                                         && Avg.             &   87.78            \\ 
    	\bottomrule[0.15em]
    \end{tabular}
    }
  }
\end{table*}

\subsection{Category knowledge reserve}
\label{sec:2.2}
Trained on enormous internet-scale corpora, LLMs are known for encoding the expert knowledge for general categories in their weights, but we ask ourselves, is the expert knowledge quintessential for FGVR already contained in MLLMs? We hypothesize that MLLMs' underperformance in FGVR tasks stems from the inadequate knowledge of subordinate-level categories. To test the hypothesis, we investigate whether LLMs utilized in MLLMs can distinguish different categories by generating discriminative descriptions. Specifically, we probe the knowledge in Idefics2 via using the prompt \texttt{["Give a brief description of distinguishing features of \{CLASS NAME\}"]}. For each subordinate-level category in Oxford-IIIT Pet-37 \citep{parkhi2012cats}, we set the number of return descriptions to 200 and then equally divided them into the train set and test set. The class names in returned descriptions are replaced with demonstrative pronouns to avoid leakage of classification labels. Similarly, we conduct linear probing experiments on top of $\hat{c}_{n}^{i}$ and $\bar{c}^{i}$. As shown in Table \ref{tab:probe_textual}, Idefics2 exhibits better classification performance than the text encoder of SigLIP, demonstrating its superiority in reserving category knowledge. Despite the rich semantics of the generated category description, the category names have lower discriminability in the representation space of Idefics2 than the text model of SigLIP, illustrated in Figure \ref{fig:vis_txt_siglip} and \ref{fig:vis_txt_Idefics2}.

\subsection{Object-category alignment} 
Since our empirical study shows that Idefics2 has an acceptable capability of object information extraction and adequate knowledge of subordinate-level categories, we hypothesize that the misalignment between the visual object and category name is the root cause. We randomly sample 120 object-category pairs from Stanford Dog-120 \citep{krause20133d}, and visualize the distributions of the last token embedding of the object $\hat{o}_{m}^{i}$ and the category name $\hat{c}_{m}^{i}$ in the same representation space. As shown in Figures \ref{fig:vis_img_txt_siglip} and \ref{fig:vis_img_txt_Idefics2}, object and category representations have significant semantic gaps. Since category names may not fully represent the semantics of the visual data \citep{lyu2024unibind}, the object cannot match the ground-truth category in the representation space and thus fails to decode into the correct category name. 

\section{Method}
\begin{figure*}[t]
    \centering
    \includegraphics[width=0.98\linewidth]{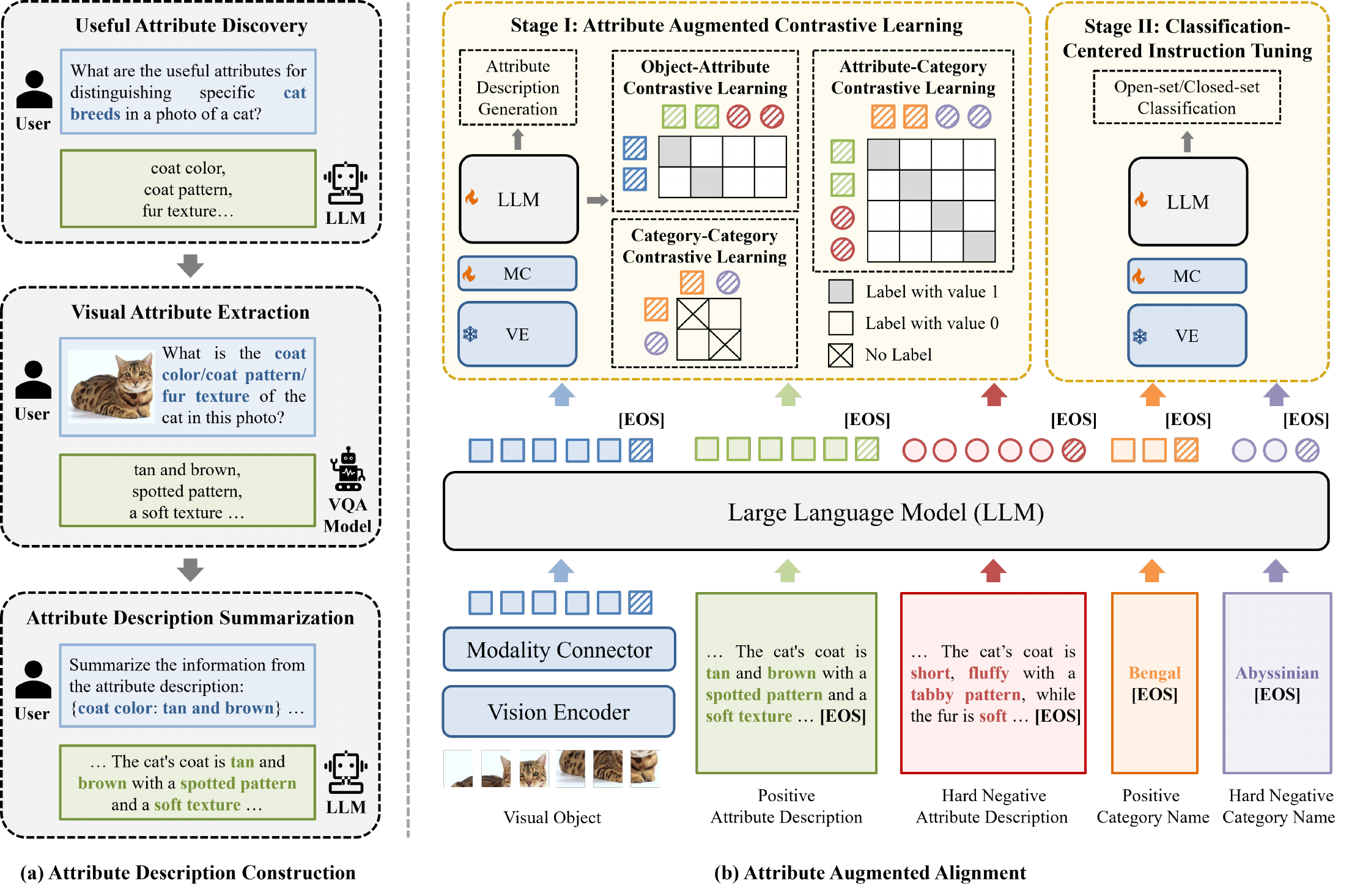}
    \caption{An illustration of framework to build \ours. (a) Attribute Description Construction, which aims to obtain informative attribute descriptions of objects. (b) Attribute Augmented Alignment, which aims to use constructed attribute descriptions to bind visual objects and category names, thus enhancing the model's FGVR capability via a two-stage training paradigm.}
    \label{fig:pipeline}
\end{figure*}

After thoroughly investigating the root cause of the underperformance in FGVR, this section formally introduces \ours, which enhances the model's FGVR performance by better aligning visual objects and category names. The framework to build \ours\ is illustrated in Figure \ref{fig:pipeline}, composed of two key components: (1) Attribute Description Construction for extracting useful attribute information that can distinguish different categories. (2) Attribute Augmented Alignment dedicated to using constructed attribute descriptions as the intermediate point to bind visual objects and category names in the representation space of LLMs, thus boosting the subsequent Classification-Centered Instruction Tuning. 

\subsection{Attribute Description Construction}

Although it has been demonstrated in \citep{liu2022universal, el2023learning} that language is a powerful tool for capturing semantic relationships, dependency on category names exclusively to align with extracted visual embeddings is unreliable. As discussed in Section \ref{sec:2.2}, leveraging adequate knowledge of subordinate-level categories, LLMs can describe distinguishing features that better capture category semantics than category names. Inspired by \citep{liu2024democratizing} that exploits a cascade of foundation models to translate useful visual information from visual to textual modality, we propose constructing sample-wise attribute descriptions for each FGVR training set. 

Specifically, the construction comprises three steps: 1) Useful Attribute Discovery by LLMs, such as GPT-4 \citep{achiam2023gpt} and LLaMA \citep{touvron2023open}. These attribute names are employed as keys to instruct Visual Question Answering (VQA) models (such as BLIP-2 \citep{li2023blip} and LLaVA \citep{liu2024visual}) for extracting useful attribute values. 2) Visual Attribute Extraction by VQA models. These attribute values enrich the information for distinguishing subordinate-level categories. 3) Attribute Description Summarization by LLMs. These descriptions help alleviate the gaps between visual objects and category names in the training phase of \ours.

\textbf{Userful Attribute Discovery.} Given the super-category of each FGVR dataset, such as \texttt{aircraft} for FGVC-Aircraft \citep{maji2013fine}, we first identify a useful set of attributes that set apart the subordinate-level categories. For example, the \texttt{wing shape} attribute can help distinguish various aircraft models. To discover such key visual cues, we tap into the expert knowledge of LLMs, which is otherwise only restricted to experts. Specifically, we ask LLMs: \texttt{["Your task is to tell me what are the useful attributes for distinguishing \{SUPERCLASS\} \{CLASSUNIT\} in a photo of a \{SUPERCLASS\}"]}. Formally, LLM takes a super-category $C_{\text {sup}}$ as input and outputs a list of useful attributes:
\begin{equation}
    N^{C_\text{sup}}=\textbf{L}_{\theta}\left(P^{\text{dis}}(C_\text{sup})\right),
\end{equation}
where $N^{C_\text{sup}} = \{N^{C_\text{sup}}_{1}, \ldots, N^{C_\text{sup}}_{s}\}$ are the generated attribute keys for the category $C_\text{sup}$, $\textbf{L}_{\theta}$ are LLM layers, and $P^{\text{dis}}$ is the How-to LLM-prompt in \citep{liu2024democratizing}. 

\textbf{Visual Attribute Extraction.} With the discovered attribute names $N^{C_\text{sup}}$, we leverage VQA models that excel at identifying general visual attributes (e.g., shape, color) of objects to extract each attribute value per sample. For example, if an attribute is \texttt{wing shape}, VQA models are prompted to give a brief description of the wing shape, which is a much easier task than recognizing many subordinate-level categories. Following \citep{liu2024democratizing}, we add a general attribute name $N^{C_\text{sup}}_{0}=\texttt {["General description of the image"]}$ and its prompt $P^\text{ext}_{0} = \texttt{["Questions: Describe this image in details. Answer:"]}$. Formally, VQA model takes as input an image $I_{i}$, its super-category $C_\text{sup}^{i}$ and the attribute names $N^{C_\text{sup}^{i}}$, the output visual attributes are given as:
\begin{equation}
    V_{i}=\mathbf{Q}_{\epsilon}\left(I_{i}, C_\text{sup}^{i}, P_{\text{ext}}(N^{C_\text{sup}^{i}})\right),
\end{equation}
where $V_{i} = \{V_{{1}}^{i}, \ldots, V_{{v}}^{i}\}$ denotes the extracted set of visual attributes for image $I_{i}$, $\textbf{Q}_{\epsilon}$ is the VQA model, and $P_\text{ext}$ is the Identify VQA-prompt in \citep{liu2024democratizing}.

\textbf{Attribute Description Construction.} After obtaining the structured set of attribute key-value pairs, we further ask LLMs: \texttt{["Summarize the information you get about the \{SUPERCLASS\} from the general description and attribute description with five sentences."]}. The summarized attribute description contains richer semantics of subordinate-level categories, making it much easier for LLM to understand. Formally, given the set of attribute names $N^{C_\text{sup}^{i}}$ and attribute values $V_{i}$, LLM outputs a summarized attribute description for image $I_{i}$:
\begin{equation}
    A_{i}=\textbf{L}_{\theta}\left(P_{\text{con}}(N^{C_\text{sup}^{i}}, V_{i})\right),
\end{equation}
where $A_{i}$ is the attribute description constructed for image $I_{i}$, and $P_\text{con}$ is the revised Reason LLM-prompt in \citep{liu2024democratizing} for summarization task only. Expanding upon our newly built attribute descriptions, we transfer traditional (object, category) pairs in FGVR datasets to (object, attribute, category) triples. Without specification, the category refers to the subordinate-level category instead of the super-category in subsequent sections. 

\subsection{Attribute Augmented Alignment}

With the constructed informative attribute descriptions, we introduce a new training paradigm named Attribute Augmented Alignment to build our \ours. It comprises two stages: (I) Attribute Augmented Contrastive Learning for aligning visual objects and category names in the representation space of LLMs. (II) Classification-Centered Instruction Tuning for enhancing the model's ability to follow the FGVR task instruction.

\textbf{Stage I: Attribute Augmented Contrastive Learning.} For each object-attribute-category triple $(O_{i}, A_{i}, C_{i})$, we utilize the vision encoder $\textbf{V}_{\alpha}$ and the learnable modality connector $\mathbf{F}_{\beta}$ to transfer $O_{i}$ into an object embedding sequence of length $S_{o}^{i}=[o_{1}^{i}, o_{2}^{i}, \dots, o_{m}^{i}]$ with length $m$. To better capture the global representations, we follow \citep{jiang2024hallucination} to pass an $\texttt{[EOS]}$ token through an embedding layer $\textbf{E}_{\phi}$ of LLM to obtain the vector representation and append it to the visual embedding sequence $S_{o}^{i}$. Therefore, we obtain the newly built object embedding sequence $\tilde{S}_{o}^{i} = [o_{1}^{i}, o_{2}^{i}, ..., o_{m}^{i}, o_{\text{EOS}}^{i}]$. Similarity, we obtain the attribute embedding sequence $\tilde{S}_{a}^{i} = [a_{1}^{i}, a_{2}^{i}, ..., a_{p}^{i}, a_{\text{EOS}}^{i}]$ with length $(p+1)$, and category embedding sequence $\tilde{S}_{c}^{i} = [c_{1}^{i}, c_{2}^{i}, ..., c_{n}^{i}, c_{\text{EOS}}^{i}]$ with length $(n+1)$. Then, $\tilde{S}_{o}^{i}, \tilde{S}_{a}^{i}, \tilde{S}_{c}^{i}$ are individually fed into LLM layers $\mathbf{L}_{\theta}$, and the embeddings of the last predicted token $\hat{o}_{\text{EOS}}^{i}, \hat{a}_{\text{EOS}}^{i}, \hat{c}_{\text{EOS}}^{i}$ are utilized as the global representations of $O_{i}, A_{i}, C_{i}$, respectively. Without specified, we use $\hat{o}^{i}=\hat{o}_{\text{EOS}}^{i}, \hat{a}^{i}=\hat{a}_{\text{EOS}}^{i}, \hat{c}^{i}=\hat{c}_{\text{EOS}}^{i}$ for simplicity.

To improve the effectiveness of contrastive learning, we then mine difficult incorrect category names for each example object $O_{i}$ used in the FGVR dataset. To do this, we use a CLIP model \citep{radford2021learning} for mining hard negative samples: for every example image, we select three images along with their attribute descriptions from the three most similar but incorrect categories. Attribute descriptions and category names from these hard negative samples are subsequently treated as additional negatives. Thus, the formulation of Object-Attribute Contrastive (OAC) loss with the inclusion of hard negatives can be described as follows:
\begin{subequations}
\begin{align}
    \mathcal{L}_{OA}^{hn}=\sum\limits_{(\hat{o}^{i}, \hat{a}^{i}, \hat{c}^{i}) \in \mathcal{B}}-\log \frac{\exp ^{Sim(\hat{o}^{i},  \hat{a}^{i})}}{\sum\limits_{\hat{a}^{j} \in \mathcal{B}} \exp ^{Sim\left(\hat{o}^{i}, \hat{a}^{j}\right)}+\sum\limits_{\hat{a}^{w} \in \mathcal{A}_{hn}^{i}} \exp ^{Sim\left(\hat{o}^{i}, \hat{a}^{w}\right)}},    
\end{align}
\begin{align}
     \mathcal{L}_{AO}=\sum\limits_{(\hat{o}^{i}, \hat{a}^{i}, \hat{c}^{i}) \in \mathcal{B}}-\log \frac{\exp ^{Sim(\hat{o}^{i}, \hat{a}^{i})}}{\sum\limits_{\hat{o}^{k} \in \mathcal{B}} \exp ^{Sim\left(\hat{o}^{k}, \hat{a}^{i}\right)}},   
\end{align}
\begin{align}
     \mathcal{L}_{OAC}^{hn}= (\mathcal{L}_{OA}^{hn} + \mathcal{L}_{AO})/2,
\end{align}
\end{subequations}
where $\mathcal{A}_{hn}^{i}$ denotes the attribute representation set of hard negatives for the object $O_{i}$, $Sim(\cdot,\cdot)$ measures the cosine similarity in a semantic space. 

Similarly, Attribute-Category Contrastive (ACC) loss with the inclusion of hard negatives is formulated as follows:
\begin{subequations}
    \begin{align}
    \mathcal{L}_{AC}^{hn}=\sum\limits_{(\hat{o}^{i}, \hat{a}^{i}, \hat{c}^{i}) \in \mathcal{B}}-\log \frac{\exp ^{Sim(\hat{a}^{i},  \hat{c}^{i})}}{\sum\limits_{\hat{c}^{j} \in \mathcal{B}} \exp ^{Sim\left(\hat{a}^{i}, \hat{c}^{j}\right)}+\sum\limits_{\hat{c}^{w} \in \mathcal{C}_{hn}^{i}} \exp ^{Sim\left(\hat{a}^{i}, \hat{c}^{w}\right)}},
    \end{align}
    \begin{align}
    \mathcal{L}_{CA}^{hn}=\sum\limits_{(\hat{o}^{i}, \hat{a}^{i}, \hat{c}^{i}) \in \mathcal{B}}-\log \frac{\exp ^{Sim(\hat{a}^{i}, \hat{c}^{i})}}{\sum\limits_{\hat{a}^{j} \in \mathcal{B}} \exp ^{Sim\left(\hat{a}^{j}, \hat{c}^{i}\right)}+\sum\limits_{\hat{a}^{w} \in \mathcal{A}_{hn}^{i}} \exp ^{Sim\left(\hat{a}^{w}, \hat{c}^{i}\right)}},        
    \end{align}
    \begin{align}
        \mathcal{L}_{ACC}^{hn} = (\mathcal{L}_{AC}^{hn} + \mathcal{L}_{CA}^{hn})/2,
    \end{align}
\end{subequations}
where $\mathcal{C}_{hn}^{i}$ denotes the category representation set of hard negatives for the object $O_{i}$. 

As discussed in Section \ref{sec:2.2}, it is hard to differentiate between category names in the representation space of LLMs. Inspired by the intra-modal contrastive loss to promote the model's ability to differentiate between hard nagative captions \citep{zhang2024contrasting}, we additionally define Category-Category Contrastive (CCC) loss as follows:
\begin{equation}
     \mathcal{L}_{CCC}=\sum\limits_{(\hat{o}^{i}, \hat{a}^{i}, \hat{c}^{i}) \in \mathcal{B}}-\log \frac{1}{\sum\limits_{\hat{c}^{k} \in \mathcal{C}_{hn}^{i}} \exp ^{Sim\left(\hat{c}^{i}, \hat{c}^{k}\right)}}.
\end{equation}
To maintain the generative power of the model, we use the attribute descriptions as LLM-augmented captiona to formulate the attribute description generation task. Therefore, the optimization object of the first stage can be defined as follows:
\begin{equation}
    \mathcal{O}_{\beta, \theta}^{\text{I}} = \mathop{\arg\min}\limits_{\beta, \theta}\mathcal{L}_{G}^{\text{att}} + (\mathcal{L}_{OAC}^{hn} + \mathcal{L}_{ACC}^{hn} + \mathcal{L}_{CCC}) / 2,
\end{equation}
where $\mathcal{L}_{G}^{\text{att}}$ denotes the attribute description generation loss.

\textbf{Stage II: Classification-Centered Instruction Tuning.} In the second stage, we formulate the FGVR dataset as two kinds of instruction tuning data: open-set QA data and closed-set multiple-choice data. Then we fine-tune the model using this classification-centered instruction tuning data. Consequently, the optimization object of the second stage can be formulated as:
\begin{equation}
    \mathcal{O}_{\beta, \theta}^{\text{II}} = \mathop{\arg\min}\limits_{\beta, \theta}\mathcal{L}_{G}^{\text{cls}},    
\end{equation}
where $\mathcal{L}_{G}^{\text{cls}}$ denotes the generation loss of classification-centered instruction tuning data.

\section{Experiments}
\begin{table*}[t]
\caption{
Comparison with leading methods on six FGVR datasets. \#P denotes parameters count. 
}
\label{tab:main_results}
    \centering
      \def\arraystretch{0.97}
     \resizebox{0.99\linewidth}{!}{
    \begin{tabular}{l|c|ccccccc}
    \toprule[0.15em]
\bf Model  & \bf  \#P & \bf Dog-120 & \bf Bird-200 & \bf Aircraft-102 & \bf Flower-102  & \bf Pet-37 & \bf Car-196 & \bf Avg. \\
\midrule 

LLaVA 1.5          & 7B     & 38.96 & 35.24 & 34.71 & 51.37 &52.25 & 46.92 & 43.24 \\
LLaVA-Next (Mistral)           & 7B   & 38.86 & 34.88 & 32.49 & 43.91 &53.72 & 49.48 & 42.22 \\
MobileVLM v2        & 7B   & 39.92 & 33.90 & 35.01 & 54.89 & 53.69 & 46.29 & 43.95 \\
InstructBLIP Vicuna  & 7B  &41.60 & 32.78 & 31.68 & 50.90 & 54.92 & 48.25 & 43.36 \\
InstructBLIP Flan-T5-XL & 4B & 47.10 & 32.15 & 29.19 & 62.29 &  59.99 & 64.58 & 49.22 \\
Phi-3-Vision          & 4B  & 39.80 & 37.63 & 42.33 & 51.59 &56.36 & 54.50 & 47.04 \\
BLIP2 Flan-T5-XL    & 4B  & 46.17 & 33.70 &32.94 & 64.32 & 65.00 & 67.68 & 51.64 \\
InternLM XComposer 2  & 7B & 41.47 & 37.42 & 40.53 & 54.25 & 63.23 & 53.89 & 48.47 \\
Pali-Gemma            & 3B & 51.68 & 36.62 & 39.87 & 69.64 &75.42 & 64.64 & 56.31 \\
Idefics1             & 9B  &  39.74 & 36.50 &  34.62 & 51.70 &  48.51 & 29.42 & 40.08 \\
Idefics2            & 8B  & 57.96 & 47.17 &  \underline{56.23} & 72.78 &  81.28 & \underline{80.25} & 65.95 \\
Qwen-VL-Chat          & 10B  &  \underline{66.18} & \underline{52.30} & 45.96 & \underline{75.95} & \underline{87.82} & 76.23 & \underline{67.41} \\
\midrule
\textbf{\ours~(ours)} & 8B & \textbf{72.86} & \textbf{57.61} & \textbf{63.82} &  \textbf{89.88} &  \textbf{92.18} & \textbf{84.67} & \textbf{76.84}  \\ 
& & \textcolor{red}{(+6.68)} & \textcolor{red}{(+5.31)} & \textcolor{red}{(+7.59)} & \textcolor{red}{(+13.93)} & \textcolor{red}{(+4.36)} & \textcolor{red}{(+4.42)} & \textcolor{red}{(+9.43)}\\
\bottomrule[0.15em]
\end{tabular}
}

\end{table*}

In this section, we evaluate the performance of \ours~aiming to answer the following questions: (1) Can \ours~effectively improve FGVR accuracy in MLLMs? (2) Does each core design of \ours~benefit the accuracy improvement? (3) Is \ours~effective in aligning visual objects and category names in the representation space of LLMs?

\subsection{Implementation Details}

\textbf{Datasets.} We conduct experiments on several popular FGVR datasets that include CaltechUCSD Bird-200 \citep{wah2011caltech}, Stanford Car-196 \citep{krause20133d}, Stanford Dog-120 \citep{krause20133d}, Flower-102 \citep{nilsback2008automated}, Oxford-IIIT Pet-37 \citep{parkhi2012cats}, and FGVC-Aircraft \citep{maji2013fine}. Following \citep{geigle2024african}, we leverage the test sets as resources for annotated data and frame FGVR as a multiple-choice task with well-defined answer candidates. To facilitate \ours, we select the training sets of them to construct attribute description, build open-set QA and closed-set multiple-choice instructing tuning data, ensuring that these images are different from the ones used in testing.

\textbf{Evaluated MLLMs.} We build \ours~upon Idefics2 \citep{laurenccon2024matters} for its open-source accessibility and leading zero-shot performance. Several recent MLLMs of comparable parameter sizes are evaluated, including LLaVA 1.5 \citep{liu2024visual}, LLaVA-Next \citep{liu2024improved}, MobileVLM v2 \citep{chu2024mobilevlm}, InstructBLIP Vicuna \citep{instructblip}, InstructBLIP Flan-T5-XL \citep{instructblip}, Phi-3-Vision \citep{abdin2024phi}, BLIP2 Flan-T5-XL \citep{li2023blip}, InternLM XComposer 2 \citep{dong2024internlm},  Pali-Gemma \footnote{\href{https://ai.google.dev/gemma/docs/paligemma/model-card}{https://ai.google.dev/gemma/docs/paligemma/model-card}}, Idefics1 \citep{laurenccon2024obelics}, Idefics2 \citep{laurenccon2024matters}, and Qwen-VL-Chat \citep{bai2023qwen}.

\textbf{Training Settings.} All seeds are fixed across the training procedures for fairness. We train \ours~using the QLoRa technique \citep{dettmers2024qlora}, updating adapters in the LLM and modality connector including perceiver resampler with 8 NVIDIA A6000 GPUs with 48G of memory. We use 4-bit quantization, with $\gamma=8$ and $\alpha=8$ for LoRa, and a learning rate of 2e-4. For both stages, the model is trained for one epoch with the warming steps of 60. The accumulated batch size is set to 64 and 128 for stage I and stage II, respectively. 

\subsection{Main Results}

In Table \ref{tab:main_results}, we compare \ours~with previous leading approaches on six popular FGVR datasets. \ours~exhibits a significantly enhanced FGVR capability compared to a wide range of MLLMs. Notably, \ours~shows superior performance than Idefics2 by an average of +10.89\% and Qwen-VL-Chat of +9.43\% across all datasets. Note that \ours~is built upon Idefics2 \citep{laurenccon2024matters}, a high-performing model in various vision-language and vision-centric tasks, and the enhanced performance on FGVR makes it a valuable foundation to benefit more advanced tasks with finer granularity.   

\begin{table*}[!t]
\caption{Analysis of \ours. "Original" represents the zero-shot performance of Idefics2.}
\begin{subtable}{.32\textwidth}
    \centering
    \caption{FT methods.}
    \label{tab:add_new_data}
    \resizebox{0.84\textwidth}{!}{%
    \begin{tabular}{l|ccc}
    \toprule[0.15em]
        \bf Method            & \bf Avg.          \\
        \midrule
        Original          & 65.95            \\
        Finetune       &   0.03          \\
        \bf \ours~(ours)   & \bf 76.84            \\
    \bottomrule[0.15em]
    \end{tabular}
    }%
\end{subtable}%
\hspace{0.83em}
\begin{subtable}{.32\textwidth}
    \centering
    \caption{Effectiveness of attributes.}
    \label{tab:attributes}
    \resizebox{0.87\textwidth}{!}{%
    \begin{tabular}{l|ccc}
    \toprule[0.15em]
        \bf Method            & \bf Avg.          \\
        \midrule
        Original          & 65.95            \\
        CL (obj.-cat.)       &   72.72          \\
        \bf CL (obj.-att.-cat.)   & \bf 76.84            \\
    \bottomrule[0.15em]
    \end{tabular}
    }%
\end{subtable}
\hspace{0.16em}
\begin{subtable}{.32\textwidth}
    \centering
    \caption{Training paradigm.}
    \label{tab:training_paradigm}
    \resizebox{0.66\textwidth}{!}{%
    \begin{tabular}{l|ccc}
    \toprule[0.15em]
        \bf Method            & \bf Avg.          \\
        \midrule
        Original          & 65.95            \\
        One stage       &   25.42          \\
        \bf Two stages   & \bf 76.84            \\
    \bottomrule[0.15em]
    \end{tabular}
    }%
\end{subtable}
\end{table*}

\subsection{Analysis of \ours}

\textbf{Does Fine-tuning Solely on Additional Open-set FGVR Data Bring Performance Gains?} In Table \ref{tab:add_new_data}, we fine-tuning Idefics2 solely on additional open-set FGVR data. We observe that it deteriorates the instruction following capability for answering multiple-choice questions in our test settings. \ours~outperforms the fine-tuned model, which indicates that \ours~effectively boosts FGVR accuracy by integrating attribute augmented alignment into the training paradigm rather than solely fine-tuning on additional data. 

\textbf{Does Attribute Descriptions Contribute Performance Gains of Contrastive Learning?} To demonstrate the impact of augmenting contrastive learning with attribute descriptions, we conduct experiments and report the results in Table \ref{tab:attributes}. In the ablation experiments, we employ contrastive learning on object-category pairs without utilizing attribute descriptions. The results show that the attribute description benefits the alignment between visual objects and category names. 

\begin{figure*}[t]
  \centering
  \begin{subfigure}[b]{0.32\linewidth}
    \includegraphics[width=\textwidth]{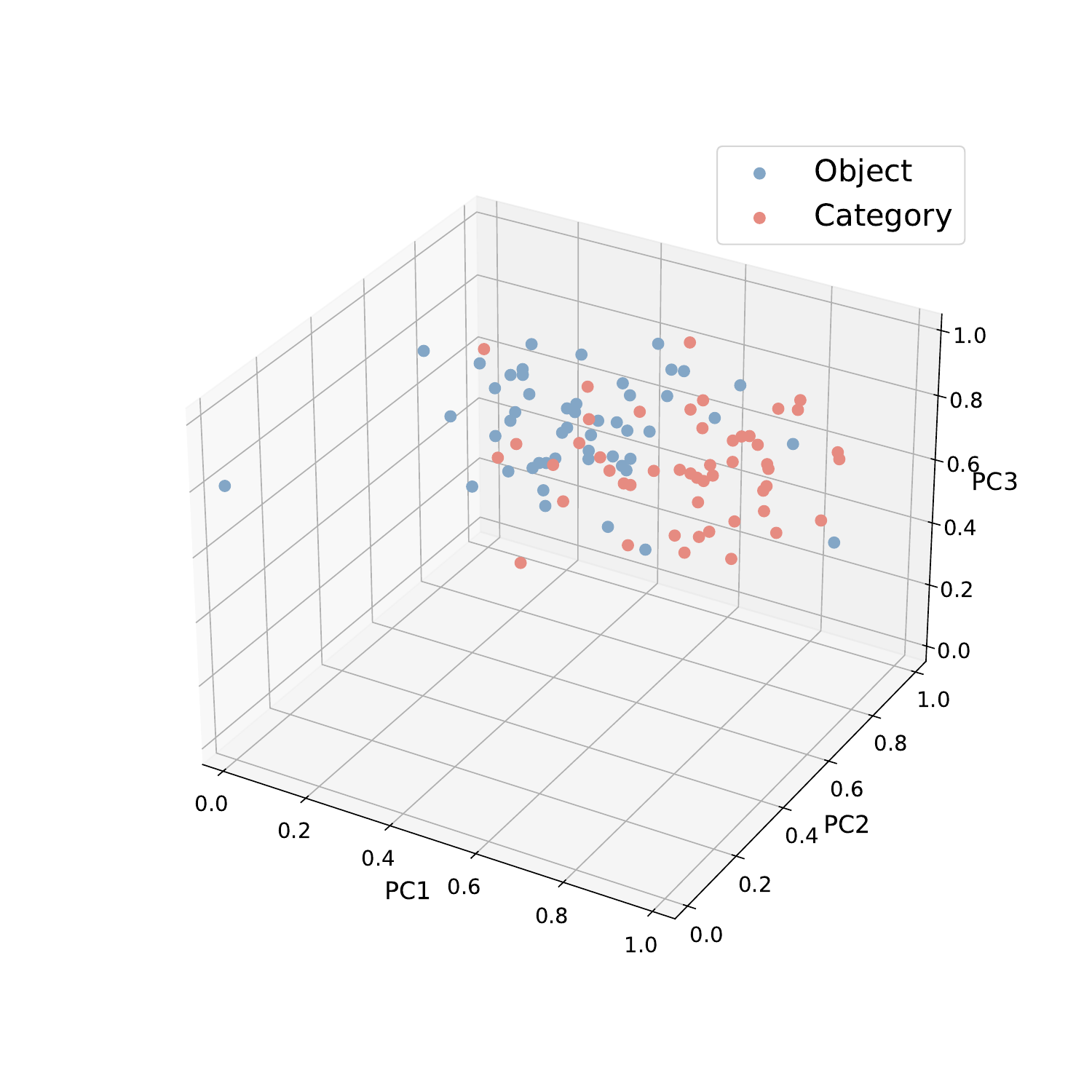}
    \vspace{-3ex}
    \caption{Finetune}
    \label{fig:FT}
  \end{subfigure}
  \hfill
  \begin{subfigure}[b]{0.32\linewidth}
    \includegraphics[width=\textwidth]{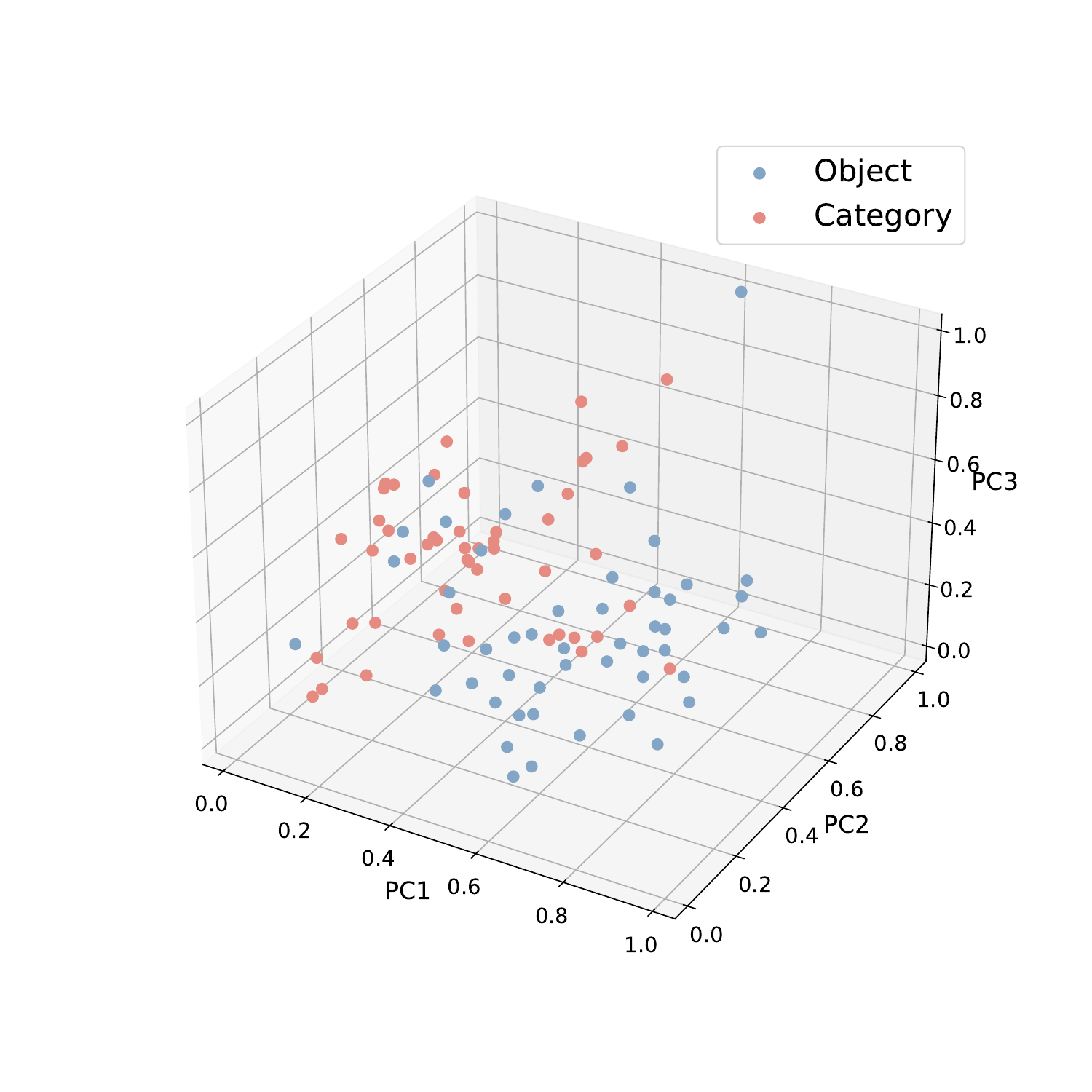}
    \vspace{-3ex}
    \caption{CL (object-category)}
    \label{fig:CL}
  \end{subfigure}
  \hfill
  \begin{subfigure}[b]{0.32\linewidth}
    \includegraphics[width=\textwidth]{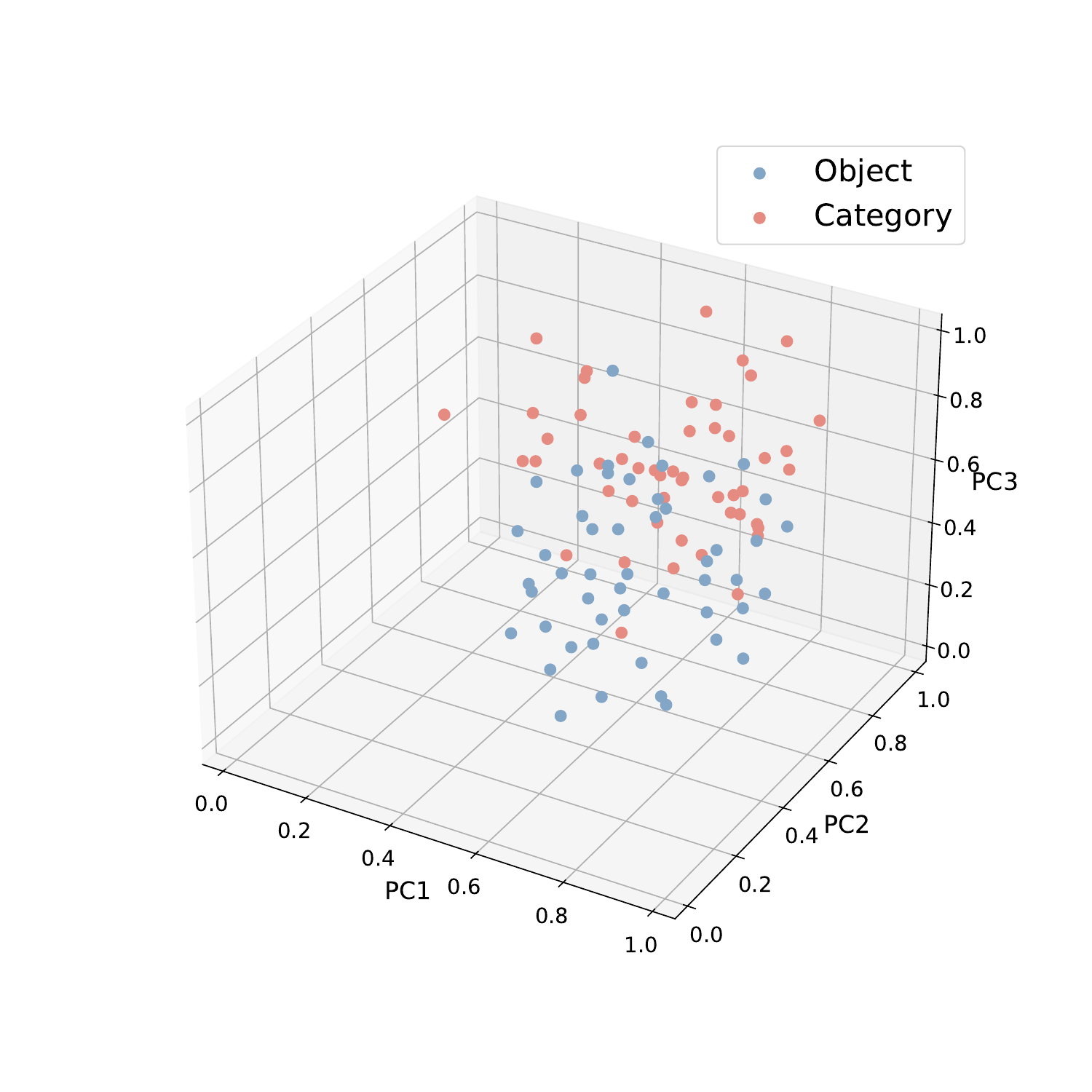}
    \vspace{-3ex}
    \caption{\ours~(ours)}
    \label{fig:ours_vis}
  \end{subfigure}
  \caption{Representation visualization of Finetune, CL (object-category) and \ours.}
  \label{fig:vis2}
\end{figure*}

\textbf{Is Training in Two Stages Necessary for Building \ours?} We further analyze the necessity of training in two stages, i.e., representation alignment before instruction tuning. Specifically, we fine-tune Idefics2 with a combined loss of classification-centered instruction tuning and attribute augmented contrastive learning. The results are reported in Table \ref{tab:training_paradigm}. We observe that fine-tuning in one stage is prone to struggle with optimization and leads to degraded performance, which indicates the effectiveness of training in two stages.

\textbf{Visualization - Does \ours~Effectively Align Visual Objects and Category Names?}  To substantiate our objective of enhancing the alignment between visual objects and category names with the auxiliary visual attributes, we randomly selected 100 data from Oxford-IIIT Pet-37 \citep{parkhi2012cats} for visualization. As illustrated in Figure \ref{fig:FT}, a substantial gap between the object and category is observable in the data distribution when fine-tuning without contrastive learning. In Figure \ref{fig:CL}, contrastive learning on object-category pairs without attribute descriptions involved fails to decrease the gap. In Figure \ref{fig:ours_vis}, with the usage of contrastive learning on object-attribute-category triples, the gap decreases significantly, thus boosting FGVR accuracy.

\section{Related Work}

\textbf{Multi-modal Large Language Models}. Multimodal Large Language Models (MLLMs) aim to enhance machines' ability to understand and process complex information by integrating multiple data modalities such as vision, text, and audio. In recent years, MLLMs have achieved significant progress in three key areas. First was large-scale pre-training and fine-tuning, as seen in models like BLIP-2 \citep{li2023blip}, LLaVA \citep{liu2024visual}, MiniGPT-4 \citep{zhu2023minigpt}, PaLM-E \citep{driess2023palm}, Kosmos-2 \citep{peng2023kosmos} and Visual ChatGPT \citep{wu2023visual}, which used pre-training on vast multimodal datasets and were then fine-tuned for specific tasks, greatly improving the models' generalization ability and task performance. The second area was cross-modal consistency, focusing on ensuring information consistency across different modalities through techniques like contrastive learning. Models such as Shikra \citep{chen2023shikra}, FROMAGe \citep{koh2023grounding}, DLP \citep{jian2024bootstrapping}, BuboGPT \citep{zhao2023bubogpt}, ChatSpot \citep{zhao2023chatspot}, and Qwen-VL \citep{bai2023qwen} enhanced performance in multimodal tasks by strengthening the alignment between modalities. The third area was interpretability and transparency. Models like ViperGPT \citep{suris2023vipergpt}, GPT-4 \citep{achiam2023gpt}, PandaGPT \citep{su2023pandagpt}, Video-LLaMA \citep{zhang2023video}, and Video-ChatGPT \citep{maaz2023video} enhanced the explainability of model decision-making by incorporating attention mechanisms and natural language feedback, enabling users to understand better and trust the model's output. Despite these achievements, MLLMs still face challenges, such as the inability to extract informative visual features, insufficient understanding of subordinate-level categories, and misalignment between visual objects and category names.

\textbf{Fine-Grained Visual Recognition}. FGVR \citep{welinder2010caltech,maji2013fine,wei2021fine} aims to classify visually similar subordinate categories under a broader super-category, often requiring expert-provided auxiliary annotations \citep{krause20133d,zhang2014part,vedaldi2014understanding,he2017fine} due to the subtle differences between objects. FGVR methods can be divided into three types: (i) attention-based methods enhance the model's ability to recognize subtle differences by focusing on the most critical areas of the image. (ii) hierarchical representation methods effectively handle the subtle differences between categories by constructing hierarchical feature representations that allow the model to refine image recognition progressively. (iii) metric learning methods improve the model's discriminative power in fine-grained classification tasks by learning a metric space where samples of the same class are closer and those of different classes are further apart. Moreover, TransHP \citep{wang2023transhp} integrated vision-language models, making FGVR less reliant on annotations and adaptable across various tasks. HI2R \citep{chen2024hypergraph} introduced a hypergraph-guided approach that captures intra-class and inter-class relationships, enhancing the model's ability to discern subtle distinctions between fine-grained categories. CLEVER \citep{choudhury2024curious} extracted non-expert descriptions from images and trained a fine-grained textual similarity model to match image descriptions with Wikipedia document sentences accurately. Recent advancements like FineR \citep{liu2024democratizing} employed large language models to translate visual attributes into text, enabling category identification without expert-defined labels.

\section{Conclusion}

In this paper, our objective is to analyze and boost the power of FGVR for MLLMs. We investigate the root cause of underperformance from three quintessential capabilities: object information extraction, category knowledge reserve, object-category alignment, we position the problem as the misalignment between visual objects and category names. To address the challenge, we propose Attribute Augmented Alignment, designed to use attribute descriptions as an intermediate point to bind them. Based on the aligned representation space, we build \ours, a new MLLM adept at identifying the subordinate-level category of the visual object. Our experiments, conducted on six popular FGVR datasets, demonstrate the remarkable performance of \ours. The validity of our methodology is substantiated through rigorous empirical studies. 

\textbf{Future Works.} While \ours~attains remarkable results across various FGVR datasets, it would encounter challenges in effectively learning new subordinate-level categories, and thus developing fine-tuning methods that can boost the continual FGVR capability for MLLMs is a promising future research direction.

\section*{ACKNOWLEDGMENTS}
This work was supported by the grants from the National Natural Science Foundation of China (61925201, 62132001, 62432001, 62373043) and Beijing Natural Science Foundation (L247006, 4252020).

\bibliographystyle{iclr2025_conference}
\bibliography{iclr2025_conference}

\appendix
\section{Appendix}
\subsection{More Ablation Studies}

\textbf{Attribute augmented alignment on other MLLMs.} 
We build \ours~upon Idefics2 \citep{laurenccon2024matters}. To confirm the general applicability of \ours, we conduct attribute augmented alignment (A$^3$) on another typical MLLM: LLaVA 1.5 \citep{liu2024visual}. As shown in Table \ref{tab:other_mllms}, after employing our proposed method, LLaVA 1.5 gains an accuracy improvement by 13.97\% on average, demonstrating the effectiveness and generalizability.

\textbf{Effects of attribute types.}
We analyze the effects of specific attribute types in FGVR tasks. Specifically, we selectively remove typical attribute types from [color, shape, texture, size] to evaluate the contribution to performance improvement. As shown in Table \ref{tab:ablation_attribute}, all four types of attributes play a crucial role in distinguishing subordinate-level categories, but the contribution varies with the dataset. For example, color and texture are more critical for specific datasets, like flowers and birds.   

\textbf{Effects of hard negatives.}
We compare using hard negatives and simple negatives for contrastive learning. Specifically, we replace hard negatives with randomly sampled simple negatives, meaning that the negatives used for contrastive learning are less visually similar to positives and easier to distinguish from them. As illustrated in Table \ref{tab:ablation_hard_negatives}, after applying contrastive learning with simple negatives, the improvement is limited. With the utilization of hard negatives, the modality gap decreases further, and the model harvests a significant accuracy improvement.

\textbf{Effects of two-stage training stages.}
We analyze the effects of two-stage training by evaluating \ours~by selectively removing specific training processes within each stage. As shown in Table \ref{tab:ablation_training_stages}, pretraining solely fails to follow the task instruction, while instruction tuning (I.T.) solely has a limited performance gain. Instead, pretraining and instruction tuning are complementary to further boost the accuracy, confirming the effectiveness of our two-stage training paradigm.

\textbf{Effects of description quality.}
We first design an empirical study to evaluate the description quality, i.e, how reliable the attribute descriptions we built. Similar to the probing experiments in Section \ref{sec:2.2}, we test the representation discriminability of our constructed attribute descriptions on the training set of Oxford-IIIT Pet-37 \cite{parkhi2012cats} with a splitting ratio of 1:1. The accuracy is 68.27\%, showing that the attribute descriptions can be well distinguished from each other though there exist subtle visual differences that are difficult to describe in words. Furthermore, to evaluate \ours's sensitivity to the description quality, we use three different quality levels of descriptions: (1) complete descriptions, (2) noisy descriptions (i.e., replacing some attribute descriptions with incorrect ones by prompting ChatGPT \cite{ChatGPT}), and (3) no descriptions (i.e., object-category alignment w/o attribute). Results in Table \ref{tab:ablation_noise} demonstrate \ours's robustness to description noise.

\textbf{Effects of levels of detailed descriptions.}
To analyze the impact of description length and detail on alignment effectiveness, we compare three different levels of detailed descriptions: (1) long descriptions, (2) short descriptions (i.e., generating short descriptions of input images with BLIP-2 \cite{li2023blip}, and (3) no descriptions (i.e., object-category alignment w/o attribute). Results in Table \ref{tab:ablation_detail} demonstrate that rich and informative category information expression plays a crucial role in boosting the alignment between visual objects and category names.

\begin{table*}[t]
\caption{
Effects of employing attribute augmented alignment (A$^3$) on LLaVA 1.5.
}
\label{tab:other_mllms}
    \centering
      \def\arraystretch{0.97}
     \resizebox{0.99\linewidth}{!}{
    \setlength{\tabcolsep}{9pt}
    \begin{tabular}{l|ccccccc}
    \toprule[0.15em]
\bf Model   & \bf Dog-120 & \bf Bird-200  & \bf Aircraft-102 & \bf Flower-102  & \bf Pet-37 & \bf Car-196 & \bf Avg. \\
\midrule 
LLaVA 1.5             & 38.96 & 35.24 & 34.71 & 51.37 &52.25 & 46.92 & 43.24 \\
LLaVA 1.5 + \textbf{A$^{3}$}   & \bf 57.10 & \bf 43.44 & \bf 44.49 & \bf 53.26 & \bf 78.50 & \bf 66.47 & \bf 57.21 \\
\bottomrule[0.15em]
\end{tabular}
}
\end{table*}

\begin{table*}[t]
\caption{
Effects of attribute types.
}
\label{tab:ablation_attribute}
    \centering
      \def\arraystretch{0.97}
     \resizebox{0.99\linewidth}{!}{
    \begin{tabular}{cccc|ccccccc}
    \toprule[0.15em]
 \bf Color & \bf Shape & \bf Texture & \bf Size & \bf Dog-120 & \bf Bird-200 & \bf Aircraft-102 & \bf Flower-102  & \bf Pet-37 & \bf Car-196 & \bf Avg.\\
\midrule 
         &\checkmark &\checkmark &\checkmark   & 71.15  &57.40&61.63&88.19&91.36   &84.22 & 75.66            \\
       \checkmark  & & \checkmark&  \checkmark   &72.49&57.75&62.02&90.19    &90.73 &\bf 85.26&76.41            \\
       \checkmark  &\checkmark & & \checkmark     &71.98&55.47&60.82&88.81   &89.48 &80.70&74.54            \\
       \checkmark   &\checkmark &\checkmark &   &  70.65 & \bf57.84 & 59.74& \bf 91.49& 90.30   & 82.94& 75.49            \\
       \checkmark & \checkmark&\checkmark &  \checkmark   & \textbf{72.86} & 57.61 & \textbf{63.82} &  89.88 &  \textbf{92.18} & 84.67 & \textbf{76.84}  \\ 

\bottomrule[0.15em]
\end{tabular}
}
\end{table*}

\begin{table*}[!t]
\caption{More ablation studies. "Original" represents the zero-shot performance of Idefics2.}
\begin{subtable}{.32\textwidth}
    \centering
    \caption{Negative types.}
    \label{tab:ablation_hard_negatives}
    \resizebox{0.99\textwidth}{!}{%
    \setlength{\tabcolsep}{10pt}
    \begin{tabular}{l|c}
    \toprule[0.15em]
       \bf Method            & \bf Avg.          \\
        \midrule
        Original          & 65.95            \\
        Simple Neg.       &   74.26          \\
        \bf Hard Neg. (ours) & \bf 76.84            \\
    \bottomrule[0.15em]
    \end{tabular}
    }%
\end{subtable}
\hspace{0.16em}
\begin{subtable}{.32\textwidth}
    \centering
    \caption{Training stages.}
    \label{tab:ablation_training_stages}
    \resizebox{0.99\textwidth}{!}{%
    \setlength{\tabcolsep}{11pt}
    \begin{tabular}{cc|c}
    \toprule[0.15em]
       \bf Pretrain & \bf I.T.            &\bf Avg.          \\
        \midrule
        \checkmark &          & 0.00            \\
          &  \checkmark    &   76.13         \\
        \checkmark & \checkmark  & \bf 76.84            \\
    \bottomrule[0.15em]
    \end{tabular}
    }%
\end{subtable}
\hspace{0.16em}
\begin{subtable}{.32\textwidth}
    \centering
    \caption{Description quality.}
    \label{tab:ablation_noise}
    \resizebox{0.99\textwidth}{!}{%
    \setlength{\tabcolsep}{11pt}
    \begin{tabular}{l|c}
    \toprule[0.15em]
       \bf Method            & \bf Avg.          \\
        \midrule
        None          & 72.72            \\
        Noisy       &   75.62          \\
        \bf Complete (ours)  & \bf 76.84            \\
    \bottomrule[0.15em]
    \end{tabular}
    }%
\end{subtable}

\vspace{1.0em}
\hspace{4em}
\begin{subtable}{.32\textwidth}
    \centering
    \caption{Levels of detailed descriptions.}
    \label{tab:ablation_detail}
    \resizebox{0.99\textwidth}{!}{%
    \setlength{\tabcolsep}{15pt}
    \begin{tabular}{l|c}
    \toprule[0.15em]
        \bf Method            & \bf Avg.          \\
        \midrule
        None          & 72.72            \\
        Short       &   76.11          \\
        \bf Long (ours)  & \bf 76.84            \\
    \bottomrule[0.15em]
    \end{tabular}
    }%
\end{subtable}
\hspace{4em}
\begin{subtable}{.32\textwidth}
    \centering
    \caption{Description construction.}
    \label{tab:ablation_upper}
    \resizebox{0.99\textwidth}{!}{%
    \setlength{\tabcolsep}{9pt}
    \begin{tabular}{l|c}
    \toprule[0.15em]
      \bf  Method           & \bf Avg.          \\
        \midrule
        Upper-bound          & 52.52            \\
        Tag-based baseline       &   50.48          \\
        \bf \ours~(ours)  & \bf 51.12            \\
    \bottomrule[0.15em]
    \end{tabular}
    }%
\end{subtable}
\end{table*}

\begin{table*}[t]
\caption{
Comparison with FineR. Clustering accuracy is used in FineR while classification accuracy is used in \ours.
}
\label{tab:finer}
    \centering
      \def\arraystretch{0.97}
     \resizebox{0.99\linewidth}{!}{
    \setlength{\tabcolsep}{12pt}
    \begin{tabular}{l|ccccccc}
    \toprule[0.15em]
\bf Model  & \bf Dog-120 & \bf Bird-200  & \bf Flower-102  & \bf Pet-37 & \bf Car-196 & \bf Avg. \\
\midrule 

FineR  & 48.10 & 51.10 & 63.80 & 72.90 & 49.20 & 57.00 \\
\textbf{\ours~(ours)} & \bf 72.86 & \bf 57.61 & \bf 89.88 & \bf 92.18 & \bf 84.67 & \bf 79.44 \\

\bottomrule[0.15em]
\end{tabular}
}
\end{table*}

\begin{table*}[t]
\caption{
Object-category alignment quality for Idefics2 and \ours.
}
\label{tab:distance}
    \centering
      \def\arraystretch{0.97}
     \resizebox{0.99\linewidth}{!}{
       \setlength{\tabcolsep}{9pt}
    \begin{tabular}{l|ccccccc}
    \toprule[0.15em]
\bf Metric  & \bf Dog-120 & \bf Bird-200 & \bf Aircraft-102 & \bf Flower-102  & \bf Pet-37 & \bf Car-196 & \bf Avg.\\
\midrule 

Idefics2  & 0.14 & 0.12  & 0.12 & 0.09 & 0.15 & 0.18 & 0.13\\
\textbf{\ours~(ours)}         &  \bf 0.28 &  \bf0.17  & \bf 0.30 &  \bf0.28 &  \bf0.28 &  \bf0.32 & \bf0.27\\
\bottomrule[0.15em]
\end{tabular}
}
\end{table*}

\begin{table*}[t]
\caption{
Comparison on common object recognition datasets.
}
\label{tab:common}
    \centering
      \def\arraystretch{0.97}
     \resizebox{0.99\linewidth}{!}{
    \setlength{\tabcolsep}{18pt}
    \begin{tabular}{l|ccccccc}
    \toprule[0.15em]
\bf Model  & \bf IN-adversarial & \bf IN-rendition  & \bf IN-sketch   & \bf Avg. \\
\midrule 

Idefics2  & \bf 79.84 &\bf 93.23 &\bf 68.21 &\bf  80.43 \\
\textbf{\ours~(ours)} & 75.96 & 92.43 & 66.77 & 78.39 \\

\bottomrule[0.15em]
\end{tabular}
}
\end{table*}

\begin{table*}[t]
\caption{
Discriminability comparison between different datasets. O and C denote visual object and category name, respectively. 
}
\label{tab:ablation_variance}
    \centering
      \def\arraystretch{0.97}
     \resizebox{0.99\linewidth}{!}{
     \setlength{\tabcolsep}{9pt}
    \begin{tabular}{l|cccccc}
    \toprule[0.15em]
\bf Metric  & \bf Dog-120 & \bf Bird-200 & \bf Aircraft-102 & \bf Flower-102  & \bf Pet-37 & \bf Car-196 \\
\midrule 

Inter-class Dist. for O ($\uparrow$)  & 0.36 &  0.37 & 0.31 & 0.41 & 0.43 & 0.49 \\
Intra-class Var. for O ($\downarrow$)         & 0.27 & 0.20  & 0.12 & 0.07 & 0.28 & 0.19 \\
Inter-class Dist. for C ($\uparrow$)          & 0.95 & 0.66  & 0.75 & 0.90 & 0.99 & 0.48 \\
\bottomrule[0.15em]
\end{tabular}
}
\end{table*}

\textbf{Comparison with FineR.}
We compare the clustering accuracy (cAcc) used in FineR \cite{liu2024democratizing} with the classification accuracy used in \ours. FineR can be considered as improving the FGVR performance by using the attributes as a zero-shot manner. Note that classification accuracy can be considered as the perfect case of Hungarian matching used to obtain clustering accuracy. Results in Table \ref{tab:finer} show the superiority of building an attribute-aware model compared to a multi-agent system using attributes in a zero-shot manner.

\textbf{Effects of attribute description construction.}
We evaluate the crucial rule of our proposed construction method acquiring per-sample attribute descriptions. To this end, we compare with an upper-bound and a baseline trained solely on Bird-200 dataset: (1) Upper-bound: we use Bird-200 \citep{wah2011caltech}’s per-sample ground-truth attributes, annotated by humans, for pretraining and instruction tuning. (2) Tag-based baseline: Given the category name, we prompt ChatGPT \citep{ChatGPT} to acquire per-class attribute descriptions for each attribute tag used in our construction process, without using actual image samples. Then, these per-class sets of attribute-description pairs are assigned to each sample belonging to the class. Since for the same super category, the attribute values can differ significantly, this class-wise attribute description construction process introduces some noise. As shown in Table \ref{tab:ablation_upper}, \ours~surpasses the tag-based baseline, demonstrating the superiority of building detailed per-sample attribute descriptions. Moreover, we can observe that the accuracy of \ours~is close to the upper-bound, showing the potential of using LLMs and VQA models to obtain large-scale attribute data for alignment without human annotations. 

\textbf{Performance on common object recognition.} We investigate how \ours~performs on common object recognition after pretraining and instruction tuning on FGVR datasets. As shown in Table \ref{tab:common}, we test on ImageNet-adversarial (IN-adversarial) \citep{adversarial}, ImageNet-rendition (IN-rendition) \citep{rendition}, and ImageNet-sketch (IN-sketch) \citep{sketch}. Results show that training solely on FGVR tasks has a minor impact on common object recognition accuracy. Therefore, training should be performed on a combined dataset comprising both coarse-grained and fine-grained recognition tasks. Developing fine-tuning methods that prevent such catastrophic forgetting to enhance FGVR capability without impacting common object recognition is a promising future research direction.

\begin{figure*}[!h]
    \centering
    \includegraphics[width=0.98\linewidth]{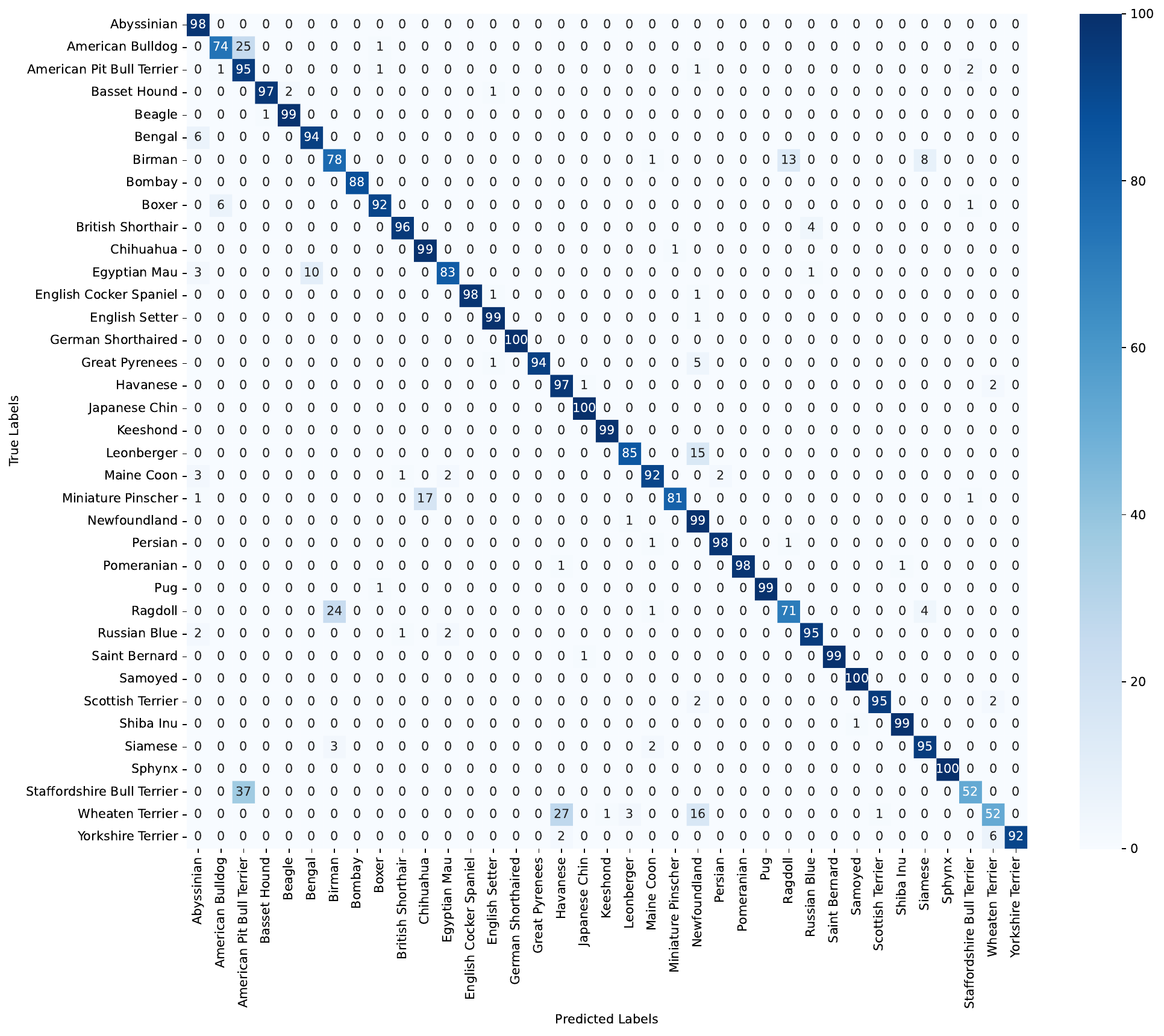}
    \caption{Confusion matrix of Oxford-IIIT Pet-37.}
    \label{fig:confusion_matrix}
\end{figure*}

\subsection{More Evaluation Metrics}
\textbf{Confusion matrix analysis.} Besides the classification accuracy, we conduct confusion matrix analysis on Oxford-IIIT Pet-37 \citep{parkhi2012cats} to identify where the model struggles across categories. As shown in Figure \ref{fig:confusion_matrix}, \ours~can recognize most of the categories correctly, while struggles in identifying a small portion of categories, such as Wheaten Terrior (52$\%$), Staffordshire Bull Terrier (58$\%$), Ragdoll (71$\%$), American Bulldog (74$\%$), and Birman (78$\%$).

\textbf{Alignment quality evaluation.} We provide a quantitative analysis of the object-category alignment quality for Idefics2 and \ours~in Table \ref{tab:distance}. The object-category alignment quality is calculated as the mean cosine similarity between embeddings of visual objects and their corresponding category names of each class. We observe that \ours~significantly increases the object-category alignment quality, which further demonstrates the effectiveness in boosting alignment.

\subsection{Discriminability comparison between different datasets}

For each dataset, we provide a quantified analysis of inter-class distance and intra-class variance for visual objects, as well as inter-class distance for category names, respectively. As shown in Table \ref{tab:ablation_variance}, Dog-120 and Aircraft-102 have the smallest inter-class distance for visual objects, meaning that the subordinate-level categories of dogs and aircrafts are typically more visually similar and difficult to distinguish. Moreover, Pet-37 and Dog-120 have the largest intra-class distance for visual objects, meaning that attribute values differ significantly for the same subordinate-level category of pets. Most importantly, category names have lower discriminability than visual objects in the representation space despite the super category. For example, though Flower-102 has large inter-class distance and small intra-class variance for visual objects, the inter-class distance for category names is not significantly different from other datasets.

\subsection{Qualitative Comparison}
\begin{figure*}[t]
    \centering
    \includegraphics[width=0.98\linewidth]{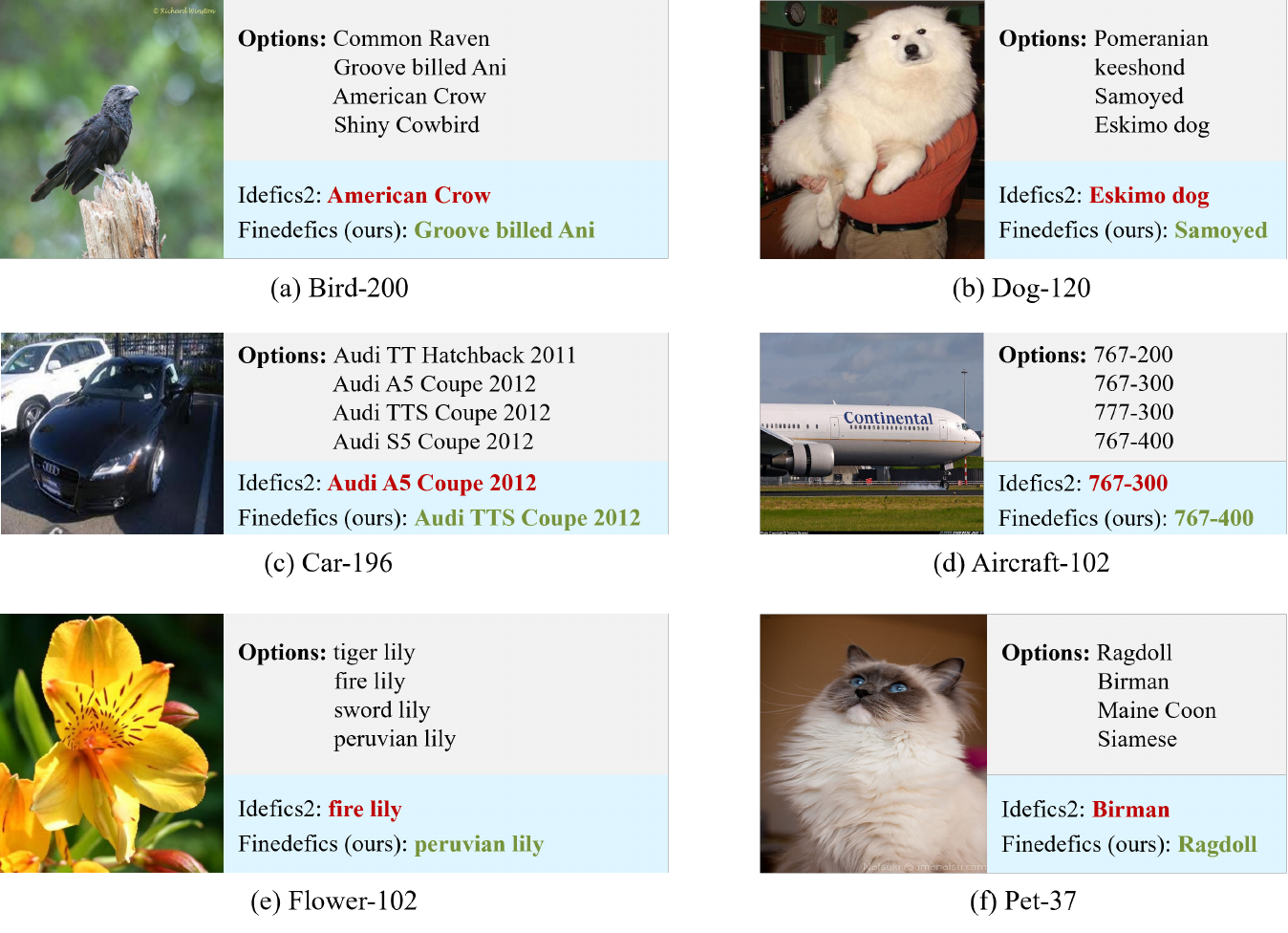}
    \caption{Qualitative comparison on FGVR datasets, where green indicates correct predictions and red indicates incorrect ones.}
    \label{fig:qualitative}
\end{figure*}

\begin{figure*}[t]
    \centering
    \includegraphics[width=0.9\linewidth]{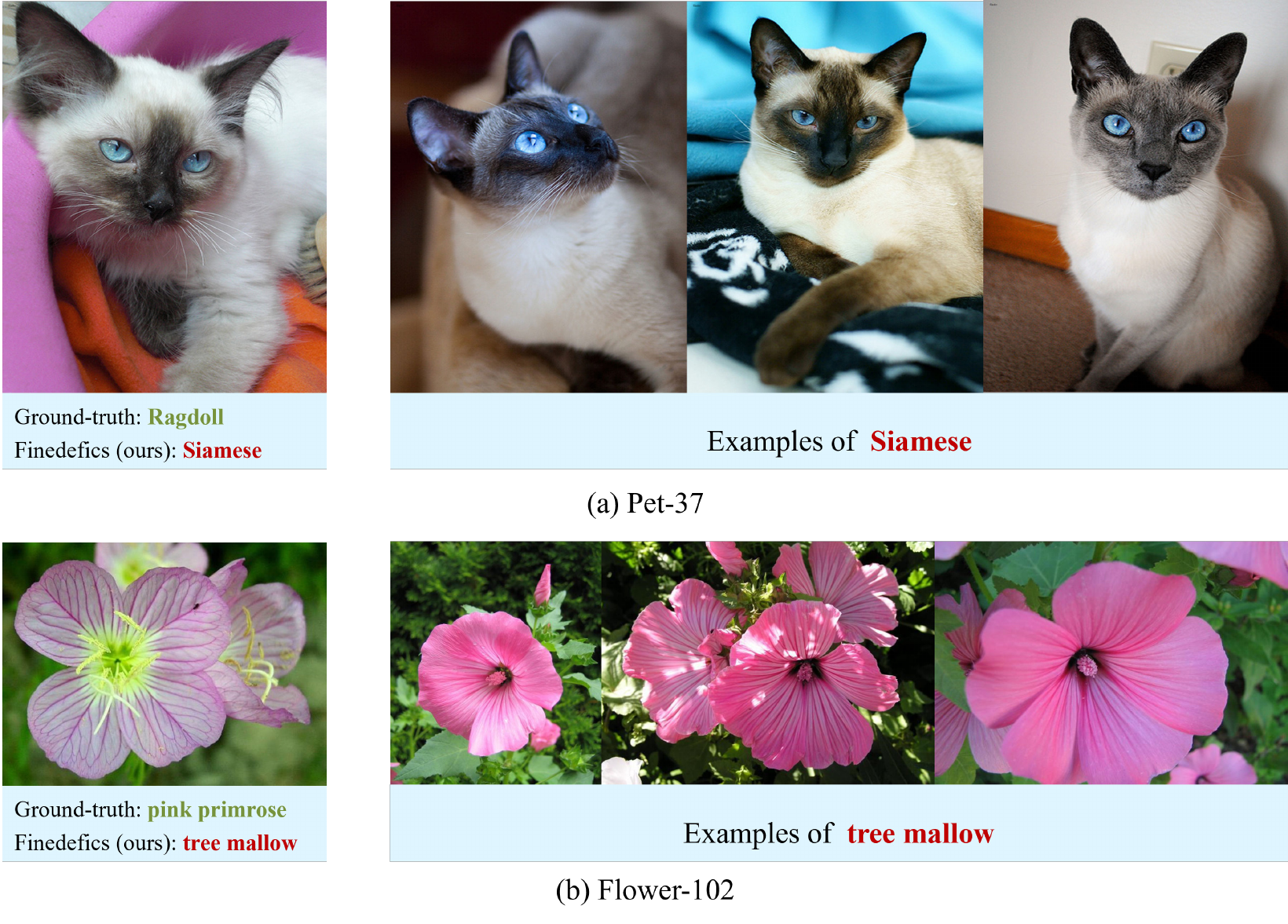}
    \caption{Error analysis examples. The left column shows the image for prediction, while the right column shows examples of the incorrect label.}
    \label{fig:error}
\end{figure*}

We visualize and analyze the predictions of \ours~and Idefics2 in Figure \ref{fig:qualitative}. \ours~successfully captures the nuance of the object features, setting them apart from visually similar sub-ordinate categories. This confirms that \ours~effectively captures fine-grained visual details from images, connects them with category names in the representation space, and then generates precise, fine-grained predictions. Furthermore, Figure \ref{fig:error} shows two examples where \ours~predicts incorrect labels. We randomly pick three examples from the incorrect label and observe that the ground-truth and predicted label share most of the attributes, with only a few exhibiting subtle differences. Concretely, as shown in the first row, Ragdolls are characterized by their fluffy, medium-to-long coats with distinct dark markings on the ears and around the eyes, whereas Siamese cats have short coats with darker color on their faces. Similarly, in the second row, pink primroses feature a light pink color with yellow-green centers, while tree mallows exhibit a vibrant pink color with dark purple to black centers.

\end{document}

%% file: math_commands.tex
\usepackage{amsmath,amsfonts,bm}

\def\eqref#1{equation~\ref{#1}}

\def\1{\bm{1}}

\DeclareMathAlphabet{\mathsfit}{\encodingdefault}{\sfdefault}{m}{sl}
\SetMathAlphabet{\mathsfit}{bold}{\encodingdefault}{\sfdefault}{bx}{n}













%% file: iclr2025_conference.bbl
\begin{thebibliography}{56}
\providecommand{\natexlab}[1]{#1}
\providecommand{\url}[1]{\texttt{#1}}
\expandafter\ifx\csname urlstyle\endcsname\relax
  \providecommand{\doi}[1]{doi: #1}\else
  \providecommand{\doi}{doi: \begingroup \urlstyle{rm}\Url}\fi

\bibitem[Abdin et~al.(2024)Abdin, Jacobs, Awan, Aneja, Awadallah, Awadalla, Bach, Bahree, Bakhtiari, Behl, et~al.]{abdin2024phi}
Marah Abdin, Sam~Ade Jacobs, Ammar~Ahmad Awan, Jyoti Aneja, Ahmed Awadallah, Hany Awadalla, Nguyen Bach, Amit Bahree, Arash Bakhtiari, Harkirat Behl, et~al.
\newblock Phi-3 technical report: A highly capable language model locally on your phone.
\newblock \emph{arXiv preprint arXiv:2404.14219}, 2024.

\bibitem[Achiam et~al.(2023)Achiam, Adler, Agarwal, Ahmad, Akkaya, Aleman, Almeida, Altenschmidt, Altman, Anadkat, et~al.]{achiam2023gpt}
Josh Achiam, Steven Adler, Sandhini Agarwal, Lama Ahmad, Ilge Akkaya, Florencia~Leoni Aleman, Diogo Almeida, Janko Altenschmidt, Sam Altman, Shyamal Anadkat, et~al.
\newblock Gpt-4 technical report.
\newblock \emph{arXiv preprint arXiv:2303.08774}, 2023.

\bibitem[Bai et~al.(2023)Bai, Bai, Yang, Wang, Tan, Wang, Lin, Zhou, and Zhou]{bai2023qwen}
Jinze Bai, Shuai Bai, Shusheng Yang, Shijie Wang, Sinan Tan, Peng Wang, Junyang Lin, Chang Zhou, and Jingren Zhou.
\newblock Qwen-vl: A frontier large vision-language model with versatile abilities.
\newblock \emph{arXiv preprint arXiv:2308.12966}, 2023.

\bibitem[Chen et~al.(2023)Chen, Zhang, Zeng, Zhang, Zhu, and Zhao]{chen2023shikra}
Keqin Chen, Zhao Zhang, Weili Zeng, Richong Zhang, Feng Zhu, and Rui Zhao.
\newblock Shikra: Unleashing multimodal llm's referential dialogue magic.
\newblock \emph{arXiv preprint arXiv:2306.15195}, 2023.

\bibitem[Chen et~al.()Chen, Wang, Li, and Yin]{chen2024hypergraph}
Lu~Chen, Qiangchang Wang, Zhaohui Li, and Yilong Yin.
\newblock Hypergraph-guided intra-and inter-category relation modeling for fine-grained visual recognition.
\newblock In \emph{ACM Multimedia 2024}.

\bibitem[Choudhury et~al.(2024)Choudhury, Laina, Rupprecht, and Vedaldi]{choudhury2024curious}
Subhabrata Choudhury, Iro Laina, Christian Rupprecht, and Andrea Vedaldi.
\newblock The curious layperson: Fine-grained image recognition without expert labels.
\newblock \emph{International Journal of Computer Vision}, 132\penalty0 (2):\penalty0 537--554, 2024.

\bibitem[Chu et~al.(2024)Chu, Qiao, Zhang, Xu, Wei, Yang, Sun, Hu, Lin, Zhang, et~al.]{chu2024mobilevlm}
Xiangxiang Chu, Limeng Qiao, Xinyu Zhang, Shuang Xu, Fei Wei, Yang Yang, Xiaofei Sun, Yiming Hu, Xinyang Lin, Bo~Zhang, et~al.
\newblock Mobilevlm v2: Faster and stronger baseline for vision language model.
\newblock \emph{arXiv preprint arXiv:2402.03766}, 2024.

\bibitem[Dai et~al.(2024)Dai, Li, Li, Tiong, Zhao, Wang, Li, Fung, and Hoi]{instructblip}
Wenliang Dai, Junnan Li, Dongxu Li, Anthony Meng~Huat Tiong, Junqi Zhao, Weisheng Wang, Boyang Li, Pascale Fung, and Steven Hoi.
\newblock Instructblip: Towards general-purpose vision-language models with instruction tuning.
\newblock \emph{arXiv preprint arXiv:2402.03766}, 2024.

\bibitem[Dettmers et~al.(2024)Dettmers, Pagnoni, Holtzman, and Zettlemoyer]{dettmers2024qlora}
Tim Dettmers, Artidoro Pagnoni, Ari Holtzman, and Luke Zettlemoyer.
\newblock Qlora: Efficient finetuning of quantized llms.
\newblock \emph{Advances in Neural Information Processing Systems}, 36, 2024.

\bibitem[Dong et~al.(2024)Dong, Zhang, Zang, Cao, Wang, Ouyang, Wei, Zhang, Duan, Cao, et~al.]{dong2024internlm}
Xiaoyi Dong, Pan Zhang, Yuhang Zang, Yuhang Cao, Bin Wang, Linke Ouyang, Xilin Wei, Songyang Zhang, Haodong Duan, Maosong Cao, et~al.
\newblock Internlm-xcomposer2: Mastering free-form text-image composition and comprehension in vision-language large model.
\newblock \emph{arXiv preprint arXiv:2401.16420}, 2024.

\bibitem[Driess et~al.(2023)Driess, Xia, Sajjadi, Lynch, Chowdhery, Ichter, Wahid, Tompson, Vuong, Yu, et~al.]{driess2023palm}
Danny Driess, Fei Xia, Mehdi~SM Sajjadi, Corey Lynch, Aakanksha Chowdhery, Brian Ichter, Ayzaan Wahid, Jonathan Tompson, Quan Vuong, Tianhe Yu, et~al.
\newblock Palm-e: An embodied multimodal language model.
\newblock \emph{arXiv preprint arXiv:2303.03378}, 2023.

\bibitem[El~Banani et~al.(2023)El~Banani, Desai, and Johnson]{el2023learning}
Mohamed El~Banani, Karan Desai, and Justin Johnson.
\newblock Learning visual representations via language-guided sampling.
\newblock In \emph{CVPR}, pp.\  19208--19220, 2023.

\bibitem[Geigle et~al.(2024)Geigle, Timofte, and Glava{\v{s}}]{geigle2024african}
Gregor Geigle, Radu Timofte, and Goran Glava{\v{s}}.
\newblock African or european swallow? benchmarking large vision-language models for fine-grained object classification.
\newblock \emph{arXiv preprint arXiv:2406.14496}, 2024.

\bibitem[He \& Peng(2017)He and Peng]{he2017fine}
Xiangteng He and Yuxin Peng.
\newblock Fine-grained image classification via combining vision and language.
\newblock In \emph{Proceedings of the IEEE conference on computer vision and pattern recognition}, pp.\  5994--6002, 2017.

\bibitem[Hendrycks et~al.(2021{\natexlab{a}})Hendrycks, Basart, Mu, Kadavath, Wang, Dorundo, Desai, Zhu, Parajuli, Guo, Song, Steinhardt, and Gilmer]{rendition}
Dan Hendrycks, Steven Basart, Norman Mu, Saurav Kadavath, Frank Wang, Evan Dorundo, Rahul Desai, Tyler Zhu, Samyak Parajuli, Mike Guo, Dawn Song, Jacob Steinhardt, and Justin Gilmer.
\newblock The many faces of robustness: A critical analysis of out-of-distribution generalization.
\newblock \emph{ICCV}, 2021{\natexlab{a}}.

\bibitem[Hendrycks et~al.(2021{\natexlab{b}})Hendrycks, Zhao, Basart, Steinhardt, and Song]{adversarial}
Dan Hendrycks, Kevin Zhao, Steven Basart, Jacob Steinhardt, and Dawn Song.
\newblock Natural adversarial examples.
\newblock \emph{CVPR}, 2021{\natexlab{b}}.

\bibitem[Jian et~al.(2024)Jian, Gao, and Vosoughi]{jian2024bootstrapping}
Yiren Jian, Chongyang Gao, and Soroush Vosoughi.
\newblock Bootstrapping vision-language learning with decoupled language pre-training.
\newblock \emph{Advances in Neural Information Processing Systems}, 36, 2024.

\bibitem[Jiang et~al.(2024)Jiang, Xu, Dong, Chen, Ye, Yan, Ye, Zhang, Huang, and Zhang]{jiang2024hallucination}
Chaoya Jiang, Haiyang Xu, Mengfan Dong, Jiaxing Chen, Wei Ye, Ming Yan, Qinghao Ye, Ji~Zhang, Fei Huang, and Shikun Zhang.
\newblock Hallucination augmented contrastive learning for multimodal large language model.
\newblock In \emph{Proceedings of the IEEE/CVF Conference on Computer Vision and Pattern Recognition}, pp.\  27036--27046, 2024.

\bibitem[Koh et~al.(2023)Koh, Salakhutdinov, and Fried]{koh2023grounding}
Jing~Yu Koh, Ruslan Salakhutdinov, and Daniel Fried.
\newblock Grounding language models to images for multimodal inputs and outputs.
\newblock In \emph{International Conference on Machine Learning}, pp.\  17283--17300. PMLR, 2023.

\bibitem[Krause et~al.(2013)Krause, Stark, Deng, and Fei-Fei]{krause20133d}
Jonathan Krause, Michael Stark, Jia Deng, and Li~Fei-Fei.
\newblock 3d object representations for fine-grained categorization.
\newblock In \emph{ICCVW}, pp.\  554--561, 2013.

\bibitem[Lauren{\c{c}}on et~al.(2024{\natexlab{a}})Lauren{\c{c}}on, Saulnier, Tronchon, Bekman, Singh, Lozhkov, Wang, Karamcheti, Rush, Kiela, et~al.]{laurenccon2024obelics}
Hugo Lauren{\c{c}}on, Lucile Saulnier, L{\'e}o Tronchon, Stas Bekman, Amanpreet Singh, Anton Lozhkov, Thomas Wang, Siddharth Karamcheti, Alexander Rush, Douwe Kiela, et~al.
\newblock Obelics: An open web-scale filtered dataset of interleaved image-text documents.
\newblock \emph{Advances in Neural Information Processing Systems}, 36, 2024{\natexlab{a}}.

\bibitem[Lauren{\c{c}}on et~al.(2024{\natexlab{b}})Lauren{\c{c}}on, Tronchon, Cord, and Sanh]{laurenccon2024matters}
Hugo Lauren{\c{c}}on, L{\'e}o Tronchon, Matthieu Cord, and Victor Sanh.
\newblock What matters when building vision-language models?
\newblock \emph{arXiv preprint arXiv:2405.02246}, 2024{\natexlab{b}}.

\bibitem[Li et~al.(2023)Li, Li, Savarese, and Hoi]{li2023blip}
Junnan Li, Dongxu Li, Silvio Savarese, and Steven Hoi.
\newblock Blip-2: Bootstrapping language-image pre-training with frozen image encoders and large language models.
\newblock In \emph{ICML}, pp.\  19730--19742. PMLR, 2023.

\bibitem[Liu et~al.(2024{\natexlab{a}})Liu, Li, Li, and Lee]{liu2024improved}
Haotian Liu, Chunyuan Li, Yuheng Li, and Yong~Jae Lee.
\newblock Improved baselines with visual instruction tuning.
\newblock In \emph{Proceedings of the IEEE/CVF Conference on Computer Vision and Pattern Recognition}, pp.\  26296--26306, 2024{\natexlab{a}}.

\bibitem[Liu et~al.(2024{\natexlab{b}})Liu, Li, Wu, and Lee]{liu2024visual}
Haotian Liu, Chunyuan Li, Qingyang Wu, and Yong~Jae Lee.
\newblock Visual instruction tuning.
\newblock \emph{Advances in neural information processing systems}, 36, 2024{\natexlab{b}}.

\bibitem[Liu et~al.(2024{\natexlab{c}})Liu, Roy, Li, Zhong, Sebe, and Ricci]{liu2024democratizing}
Mingxuan Liu, Subhankar Roy, Wenjing Li, Zhun Zhong, Nicu Sebe, and Elisa Ricci.
\newblock Democratizing fine-grained visual recognition with large language models.
\newblock \emph{arXiv preprint arXiv:2401.13837}, 2024{\natexlab{c}}.

\bibitem[Liu et~al.(2022)Liu, Xiong, Lv, Liu, and Yu]{liu2022universal}
Zhenghao Liu, Chenyan Xiong, Yuanhuiyi Lv, Zhiyuan Liu, and Ge~Yu.
\newblock Universal vision-language dense retrieval: Learning a unified representation space for multi-modal retrieval.
\newblock \emph{arXiv preprint arXiv:2209.00179}, 2022.

\bibitem[Lyu et~al.(2024)Lyu, Zheng, Zhou, and Wang]{lyu2024unibind}
Yuanhuiyi Lyu, Xu~Zheng, Jiazhou Zhou, and Lin Wang.
\newblock Unibind: Llm-augmented unified and balanced representation space to bind them all.
\newblock In \emph{CVPR}, pp.\  26752--26762, 2024.

\bibitem[Maaz et~al.(2023)Maaz, Rasheed, Khan, and Khan]{maaz2023video}
Muhammad Maaz, Hanoona Rasheed, Salman Khan, and Fahad~Shahbaz Khan.
\newblock Video-chatgpt: Towards detailed video understanding via large vision and language models.
\newblock \emph{arXiv preprint arXiv:2306.05424}, 2023.

\bibitem[Maji et~al.(2013)Maji, Rahtu, Kannala, Blaschko, and Vedaldi]{maji2013fine}
Subhransu Maji, Esa Rahtu, Juho Kannala, Matthew Blaschko, and Andrea Vedaldi.
\newblock Fine-grained visual classification of aircraft.
\newblock \emph{arXiv preprint arXiv:1306.5151}, 2013.

\bibitem[Nilsback \& Zisserman(2008)Nilsback and Zisserman]{nilsback2008automated}
Maria-Elena Nilsback and Andrew Zisserman.
\newblock Automated flower classification over a large number of classes.
\newblock In \emph{ICVGIP}, pp.\  722--729. IEEE, 2008.

\bibitem[OpenAI(2023)]{ChatGPT}
OpenAI.
\newblock Chatgpt.
\newblock \url{https://openai.com/blog/chatgpt/}, 2023.

\bibitem[Parkhi et~al.(2012)Parkhi, Vedaldi, Zisserman, and Jawahar]{parkhi2012cats}
Omkar~M Parkhi, Andrea Vedaldi, Andrew Zisserman, and CV~Jawahar.
\newblock Cats and dogs.
\newblock In \emph{CVPR}, pp.\  3498--3505. IEEE, 2012.

\bibitem[Peng et~al.(2023)Peng, Wang, Dong, Hao, Huang, Ma, and Wei]{peng2023kosmos}
Zhiliang Peng, Wenhui Wang, Li~Dong, Yaru Hao, Shaohan Huang, Shuming Ma, and Furu Wei.
\newblock Kosmos-2: Grounding multimodal large language models to the world.
\newblock \emph{arXiv preprint arXiv:2306.14824}, 2023.

\bibitem[Radford et~al.(2021)Radford, Kim, Hallacy, Ramesh, Goh, Agarwal, Sastry, Askell, Mishkin, Clark, et~al.]{radford2021learning}
Alec Radford, Jong~Wook Kim, Chris Hallacy, Aditya Ramesh, Gabriel Goh, Sandhini Agarwal, Girish Sastry, Amanda Askell, Pamela Mishkin, Jack Clark, et~al.
\newblock Learning transferable visual models from natural language supervision.
\newblock In \emph{ICML}, pp.\  8748--8763. PMLR, 2021.

\bibitem[Su et~al.(2023)Su, Lan, Li, Xu, Wang, and Cai]{su2023pandagpt}
Yixuan Su, Tian Lan, Huayang Li, Jialu Xu, Yan Wang, and Deng Cai.
\newblock Pandagpt: One model to instruction-follow them all.
\newblock \emph{arXiv preprint arXiv:2305.16355}, 2023.

\bibitem[Sur{\'\i}s et~al.(2023)Sur{\'\i}s, Menon, and Vondrick]{suris2023vipergpt}
D{\'\i}dac Sur{\'\i}s, Sachit Menon, and Carl Vondrick.
\newblock Vipergpt: Visual inference via python execution for reasoning.
\newblock In \emph{Proceedings of the IEEE/CVF International Conference on Computer Vision}, pp.\  11888--11898, 2023.

\bibitem[Touvron et~al.(2023)Touvron, Lavril, Izacard, Martinet, Lachaux, Lacroix, Rozi{\`e}re, Goyal, Hambro, Azhar, et~al.]{touvron2023open}
H~Touvron, T~Lavril, G~Izacard, X~Martinet, MA~Lachaux, T~Lacroix, B~Rozi{\`e}re, N~Goyal, E~Hambro, F~Azhar, et~al.
\newblock Open and efficient foundation language models.
\newblock \emph{Preprint at arXiv. https://doi. org/10.48550/arXiv}, 2302, 2023.

\bibitem[Van~der Maaten \& Hinton(2008)Van~der Maaten and Hinton]{van2008visualizing}
Laurens Van~der Maaten and Geoffrey Hinton.
\newblock Visualizing data using t-sne.
\newblock \emph{Journal of machine learning research}, 9\penalty0 (11), 2008.

\bibitem[Vedaldi et~al.(2014)Vedaldi, Mahendran, Tsogkas, Maji, Girshick, Kannala, Rahtu, Kokkinos, Blaschko, Weiss, et~al.]{vedaldi2014understanding}
Andrea Vedaldi, Siddharth Mahendran, Stavros Tsogkas, Subhransu Maji, Ross Girshick, Juho Kannala, Esa Rahtu, Iasonas Kokkinos, Matthew~B Blaschko, David Weiss, et~al.
\newblock Understanding objects in detail with fine-grained attributes.
\newblock In \emph{Proceedings of the IEEE conference on computer vision and pattern recognition}, pp.\  3622--3629, 2014.

\bibitem[Wah et~al.(2011)Wah, Branson, Welinder, Perona, and Belongie]{wah2011caltech}
Catherine Wah, Steve Branson, Peter Welinder, Pietro Perona, and Serge Belongie.
\newblock The caltech-ucsd birds-200-2011 dataset.
\newblock 2011.

\bibitem[Wang et~al.(2019)Wang, Ge, Lipton, and Xing]{sketch}
Haohan Wang, Songwei Ge, Zachary Lipton, and Eric~P Xing.
\newblock Learning robust global representations by penalizing local predictive power.
\newblock In \emph{NeurIPS}, pp.\  10506--10518, 2019.

\bibitem[Wang et~al.(2023)Wang, Sun, Li, and Yang]{wang2023transhp}
Wenhao Wang, Yifan Sun, Wei Li, and Yi~Yang.
\newblock Transhp: Image classification with hierarchical prompting.
\newblock \emph{Advances in Neural Information Processing Systems}, 36:\penalty0 28187--28200, 2023.

\bibitem[Wei et~al.(2021)Wei, Song, Mac~Aodha, Wu, Peng, Tang, Yang, and Belongie]{wei2021fine}
Xiu-Shen Wei, Yi-Zhe Song, Oisin Mac~Aodha, Jianxin Wu, Yuxin Peng, Jinhui Tang, Jian Yang, and Serge Belongie.
\newblock Fine-grained image analysis with deep learning: A survey.
\newblock \emph{TPAMI}, 44\penalty0 (12):\penalty0 8927--8948, 2021.

\bibitem[Welinder et~al.(2010)Welinder, Branson, Mita, Wah, Schroff, Belongie, and Perona]{welinder2010caltech}
Peter Welinder, Steve Branson, Takeshi Mita, Catherine Wah, Florian Schroff, Serge Belongie, and Pietro Perona.
\newblock Caltech-ucsd birds 200.
\newblock 2010.

\bibitem[Wu et~al.(2023)Wu, Yin, Qi, Wang, Tang, and Duan]{wu2023visual}
Chenfei Wu, Shengming Yin, Weizhen Qi, Xiaodong Wang, Zecheng Tang, and Nan Duan.
\newblock Visual chatgpt: Talking, drawing and editing with visual foundation models.
\newblock \emph{arXiv preprint arXiv:2303.04671}, 2023.

\bibitem[Zhai et~al.()Zhai, Mustafa, Kolesnikov, and Beyer]{zhai2023sigmoid}
Xiaohua Zhai, Basil Mustafa, Alexander Kolesnikov, and Lucas Beyer.
\newblock Sigmoid loss for language image pre-training.

\bibitem[Zhang et~al.(2023{\natexlab{a}})Zhang, Li, and Bing]{zhang2023video}
Hang Zhang, Xin Li, and Lidong Bing.
\newblock Video-llama: An instruction-tuned audio-visual language model for video understanding.
\newblock \emph{arXiv preprint arXiv:2306.02858}, 2023{\natexlab{a}}.

\bibitem[Zhang et~al.(2024{\natexlab{a}})Zhang, Awal, and Agrawal]{zhang2024contrasting}
Le~Zhang, Rabiul Awal, and Aishwarya Agrawal.
\newblock Contrasting intra-modal and ranking cross-modal hard negatives to enhance visio-linguistic compositional understanding.
\newblock In \emph{CVPR}, pp.\  13774--13784, 2024{\natexlab{a}}.

\bibitem[Zhang et~al.(2014)Zhang, Donahue, Girshick, and Darrell]{zhang2014part}
Ning Zhang, Jeff Donahue, Ross Girshick, and Trevor Darrell.
\newblock Part-based r-cnns for fine-grained category detection.
\newblock In \emph{Computer Vision--ECCV 2014: 13th European Conference, Zurich, Switzerland, September 6-12, 2014, Proceedings, Part I 13}, pp.\  834--849. Springer, 2014.

\bibitem[Zhang et~al.(2023{\natexlab{b}})Zhang, Wang, Cao, Xu, Ouyang, Zhao, Ding, Zhang, Duan, Yan, et~al.]{zhang2023internlm}
Pan Zhang, Xiaoyi Dong~Bin Wang, Yuhang Cao, Chao Xu, Linke Ouyang, Zhiyuan Zhao, Shuangrui Ding, Songyang Zhang, Haodong Duan, Hang Yan, et~al.
\newblock Internlm-xcomposer: A vision-language large model for advanced text-image comprehension and composition.
\newblock \emph{arXiv preprint arXiv:2309.15112}, 2023{\natexlab{b}}.

\bibitem[Zhang et~al.(2024{\natexlab{b}})Zhang, Unell, Wang, Ghosh, Su, Schmidt, and Yeung-Levy]{zhang2024visually}
Yuhui Zhang, Alyssa Unell, Xiaohan Wang, Dhruba Ghosh, Yuchang Su, Ludwig Schmidt, and Serena Yeung-Levy.
\newblock Why are visually-grounded language models bad at image classification?
\newblock \emph{arXiv preprint arXiv:2405.18415}, 2024{\natexlab{b}}.

\bibitem[Zhao et~al.(2023{\natexlab{a}})Zhao, Yu, Ge, Yang, Wei, Zhou, Sun, Peng, Dong, Han, et~al.]{zhao2023chatspot}
Liang Zhao, En~Yu, Zheng Ge, Jinrong Yang, Haoran Wei, Hongyu Zhou, Jianjian Sun, Yuang Peng, Runpei Dong, Chunrui Han, et~al.
\newblock Chatspot: Bootstrapping multimodal llms via precise referring instruction tuning.
\newblock \emph{arXiv preprint arXiv:2307.09474}, 2023{\natexlab{a}}.

\bibitem[Zhao et~al.(2023{\natexlab{b}})Zhao, Lin, Zhou, Huang, Feng, and Kang]{zhao2023bubogpt}
Yang Zhao, Zhijie Lin, Daquan Zhou, Zilong Huang, Jiashi Feng, and Bingyi Kang.
\newblock Bubogpt: Enabling visual grounding in multi-modal llms.
\newblock \emph{arXiv preprint arXiv:2307.08581}, 2023{\natexlab{b}}.

\bibitem[Zhou et~al.(2023)Zhou, Cui, Yoon, Zhang, Deng, Finn, Bansal, and Yao]{zhou2023analyzing}
Yiyang Zhou, Chenhang Cui, Jaehong Yoon, Linjun Zhang, Zhun Deng, Chelsea Finn, Mohit Bansal, and Huaxiu Yao.
\newblock Analyzing and mitigating object hallucination in large vision-language models.
\newblock \emph{arXiv preprint arXiv:2310.00754}, 2023.

\bibitem[Zhu et~al.(2023)Zhu, Chen, Shen, Li, and Elhoseiny]{zhu2023minigpt}
Deyao Zhu, Jun Chen, Xiaoqian Shen, Xiang Li, and Mohamed Elhoseiny.
\newblock Minigpt-4: Enhancing vision-language understanding with advanced large language models.
\newblock \emph{arXiv preprint arXiv:2304.10592}, 2023.

\end{thebibliography}
